\PassOptionsToPackage{prologue,dvipsnames}{xcolor}
\documentclass[10pt,twocolumn,letterpaper]{article}
\usepackage[accsupp]{axessibility}

\usepackage[pagenumbers]{cvpr} % To force page numbers, e.g. for an arXiv version
\usepackage{tabu}               
\usepackage{multirow}
\usepackage{adjustbox}

\usepackage{graphicx}
\usepackage{xcolor}
\usepackage{tabularray}

\usepackage{scrextend}
\deffootnote[.15in]{.5in}{.195in}{\makebox[.5in][r]{\thefootnotemark\hspace{.13in}}}

\usepackage[dvipsnames]{xcolor}

\definecolor{cvprblue}{rgb}{0.21,0.49,0.74}
\usepackage[pagebackref,breaklinks,colorlinks,citecolor=cvprblue]{hyperref}

\def\confName{CVPR}
\def\confYear{2024}

\title{Low-Latency Neural Stereo Streaming}

\author{Qiqi Hou \qquad Farzad Farhadzadeh \qquad   Amir Said \qquad  Guillaume Sautiere \qquad  Hoang Le$^{*}$ \\
Qualcomm AI Research$^{\dagger}$\\
{\tt\small \{qhou, ffarhadz, asaid, gsautie, hoanle\}@qti.qualcomm.com}
}

\begin{document}
\maketitle

\renewcommand{\thefootnote}{\fnsymbol{footnote}}
\footnotetext[1]{\hspace{-0.1in}Corresponding author} 
\footnotetext[2]{\hspace{-0.1in}Qualcomm AI Research is an initiative of Qualcomm Technologies, Inc.} 
\renewcommand*{\thefootnote}{\arabic{footnote}}

\begin{abstract}

The rise of new video modalities like virtual reality or autonomous driving has increased the demand for efficient multi-view video compression methods, both in terms of rate-distortion (R-D) performance and in terms of delay and runtime. While most recent stereo video compression approaches have shown promising performance, they compress left and right views sequentially, leading to poor parallelization and runtime performance. This work presents Low-Latency neural codec for Stereo video Streaming (LLSS), a novel parallel stereo video coding method designed for fast and efficient low-latency stereo video streaming.  Instead of using a sequential cross-view motion compensation like existing methods, LLSS introduces a bidirectional feature shifting module to directly exploit mutual information among views and encode them effectively with a joint cross-view prior model for entropy coding. Thanks to this design, LLSS processes left and right views in parallel, minimizing latency; all while substantially improving R-D performance compared to both existing neural and conventional codecs.

\end{abstract}    

% \vspace{-0.1in}
\section{Introduction}
% \vspace{-0.05in}
\label{sec:intro}
\begin{figure}[thp]
    \newlength\indentvspace
    \setlength{\indentvspace}{1.4mm}
        \begin{tabular}{cc}
        \hspace{-5mm}
        % \begin{adjustbox}{valign=t}
        \begin{adjustbox}{valign=t}
        \footnotesize
        % \vspace{0.5in}
        {\renewcommand{\arraystretch}{2.5}
            \begin{tabular}{r}
                \vspace{\indentvspace} HEVC~\cite{hevc} \vspace{\indentvspace}\\
                \vspace{\indentvspace} MV-HEVC~\cite{mvhevc} \vspace{\indentvspace} \\
                \vspace{\indentvspace} LSVC~\cite{Chen2022-xe} \vspace{\indentvspace}\\
                \vspace{\indentvspace} \textbf{LLSS} \vspace{\indentvspace}
            \end{tabular}
        }
        \end{adjustbox}
        \hspace{-6mm}
        &
        \begin{adjustbox}{valign=t}
        \tiny
            \begin{tabular}{c}
                \includegraphics[width=0.3\textwidth]{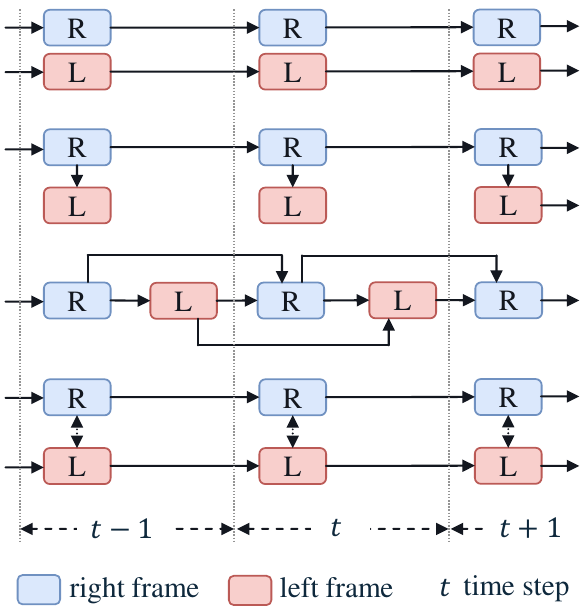}
            \end{tabular}
        \end{adjustbox} 
        \\
    \end{tabular}\vspace{-0.15in}
    \caption{
        Comparison of multi-view compression strategies. In contrast to LSVC~\cite{Chen2022-xe}, our approach processes the left and right frames simultaneously. This parallel processing not only facilitates more rate-efficient coding, it also reduces the latency between the left and right views.
    }
    \label{fig:teaser}\vspace{-0.25in}

\end{figure}

The rise in popularity of autonomous vehicles (AVs) equipped with stereo cameras, along with the widespread use of virtual reality (VR) headsets, has led to a significant increase in stereo video data. For AVs, stereo cameras serve as a cost-effective alternative to sensors like LIDAR or RADAR. The data they capture is crucial for time-sensitive safety analyses during vehicle operation, necessitating low-latency data transmission. In VR, to achieve an immersive user experience, the demands for both resolution and latency are even higher. For both AV and VR applications, it's crucial that the codec encodes stereo video efficiently while maintaining low latency.

A basic approach to stereo video coding would apply a low-delay, single-view codec like AVC~\cite{wiegand2003overview} or HEVC~\cite{hevc} to each view independently. While these traditional codecs yield promising results, and are used in commercial products, such as Meta Quest~\cite{meta_quest_2022}, they double the rate and ignore the similarities between two views. Consequently, several standard codecs have been proposed to reduce the redundancy between two views through \emph{disparity compensation} \cite{Vetro2011,Tech2016}. They typically first encode the right view frame using a single-view codec. Then left view frame is predicted from the encoded right view frame. However, these sequential processing limits the ability to process multiple views simultaneously.

Recent years have seen rapid progress in single-view neural video codecs \cite{agustsson2020ssf,le2022mobilecodec,le2022gamecodec,zhihao2022c2f,hu2021fvc,lu2019dvc,rippel2021elfvc,li2023neural,pourreza2023boosting}, particularly in low-delay settings. For instance, recent DCVC-DC~\citet{li2023neural} has shown better Rate-Distortion (R-D) performance than the H.266 standard codec~\cite{bross2021vvc}. Our method is inspired by the most recent work LSVC~\cite{Chen2022-xe}, which is the first neural codec for stereo video. Although LSVC achieves great results and shows significant superiority over MV-HEVC~\cite{mvhevc} codec, it sequentially processes the right and left view frames, which constrains its suitability for low-latency applications like VR and AVs.

In this work, we present a Low-Latency Stereo video Streaming (LLSS) codec designed for parallel stereo video coding.  This codec's development is grounded in two key insights. 
First, inspired from the recent progresses in the stereo matching methods~\cite{shen2022pcw,guo2019group,chang2018pyramid}, the \emph{disparity compensation} module between left and right views can be greatly simplified, compared to complex motion compensation schemes in LSVC \cite{Chen2022-xe}. It can be efficiently represented with horizontal shifts. Second, we observed that these disparity compensations can be executed concurrently for both views. A careful encoder design, sharing horizontally-shifted features across views, can implicitly estimate disparity, while facilitating parallel processing of both views, thereby achieving low-latency inference. Figure \ref{fig:teaser} shows a schematic comparison of these approaches. We introduce a novel component, \emph{BiShiftMod} (Bidirectional Shifting Module), which facilitates the connections and information exchange between views in our network. This module is integrated into both the codec and hypercodec~\cite{balle2018variational,minnen2018joint}, which enables data-dependent optimization of the cross-view mutual information. 
By following this approach, we replace the sequential disparity compensation with a parallel coding network that can exploit cross-view mutual information in a ``disparity-agnostic'' fashion.

Finally, we show that our solution substantially improves R-D performance compared to the state-of-the-art method on three common stereo video benchmarks, with 50.6\% BD-rate savings on the CityScapes dataset~\cite{Cordts2016-de}, 18.2\% on the KITTI 2012 dataset~\cite{Geiger2012-cy} and 15.8\% on the KITTI 2015 dataset~\cite{Menze2015-rw}. Besides, we also provide a neural network complexity and inference time study, and show that our model has only 35\% of the complexity of LSVC~\cite{Chen2022-xe} in terms of FLOPS. We further ablate each design choice to showcase the contribution of the proposed modules toward the final R-D performance.

The contributions of this paper include:
\begin{itemize}
    \item A novel low-latency neural stereo video codec architecture that replaces sequential inter-view compensation with an efficient and parallelizable learned module to connect parallel autoencoders
    \item A bidirectional shift module that effectively captures and exhibits redundancy between inter-view features
    \item A set of thorough experiments demonstrating that our method is fast, efficient, and obtaining comparable and often better than state-of-the-art methods 
\end{itemize}

% \vspace{-0.1in}
\section{Related Work}
% \vspace{-0.05in}
\label{sec:related}
\subsection{Neural video codecs}
Neural networks have been successfully applied to data compression in many domains, including the image~\cite{theis2017lossy,toderici2017full,balle2018variational,minnen2018joint,agustsson2019extreme,mentzer2020hific,he2022elic,agustsson2022multirealism,muckley2023improving,ghouse2023residual} and video~\cite{le2022mobilecodec,le2022gamecodec,habibian2019video,rippel2019lvc,agustsson2020ssf,zhihao2022c2f,hu2021fvc,lu2019dvc,rippel2021elfvc,li2021deep,sheng2022temporal,li2022hybrid,li2023neural} settings. Most of these lossy compression systems are composed of one or more variational autoencoders referred to as \emph{compressive autoencoders}\cite{kingma2013autoencoding,theis2017lossy,habibian2019video}. Lossy compression is achieved through quantization of the latent variables in the bottleneck. These latents are further compressed in a lossless manner via entropy coding, typically using a learned, data-dependent prior model.

\begin{figure*}
\centering
  \includegraphics[width=1.0\textwidth]{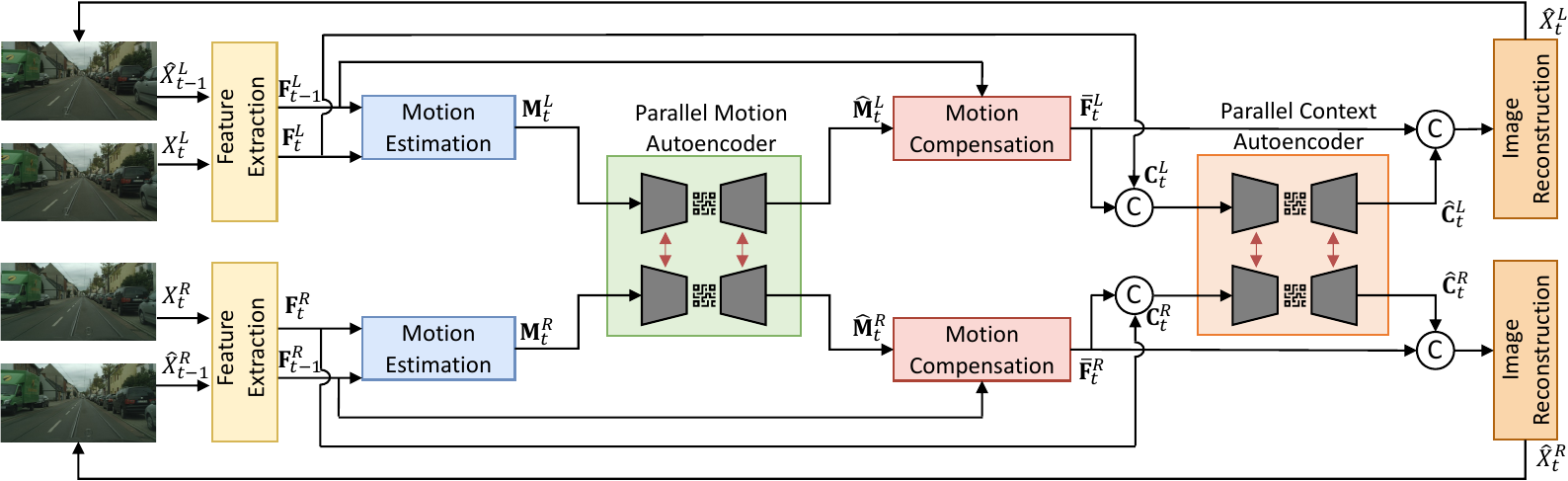} \vspace{-0.20in}
  \caption{Overall architecture of our network. It contains two branches dedicated to processing the left and right view. It incorporates a parallel motion autoencoder and a parallel context autoencoder to reduce the redundant motion and context information across views, respectively. The weights are shared across views, including the feature extraction module, the motion estimation module,  the motion compensation module, and the image reconstruction module. 
  } \vspace{-0.20in}
  \label{fig:network}
\end{figure*}

The main advantage of neural codecs is that they learn to compress from example data, whereas handcrafted codecs require expert design.  This allows for easy customization to new domains~\cite{habibian2019video}, or even to specific videos or datapoints~\cite{strumpler2021implicit,rozendaal2021instance,rozendaal2021overfitting, zhang2021implicit}.  Additionally, they may provide advantages from a deployment perspective. In practice, standard codecs often use hardware-based implementations to enable efficient operation, especially on mobile devices. However, these implementations tend to require a longer deployment process. In contrast, software-based neural codecs only need generic and ubiquitous AI accelerators for operation, making them more flexible and with the potential to enhance various application domains, especially where hardware-based codecs are not available. Lastly, neural codecs can be optimized end-to-end to improve \emph{perceptual quality} through the use of perceptual loss functions \cite{yang2021perceptual, agustsson2019extreme, mentzer2020hific, muckley2023improving, ghouse2023residual}, or take the semantics of the video into account via region-of-interest coding \cite{cai2019end, fathima2023spatialratedistortion}.   Despite these dissimilarities, neural video codecs have taken inspiration from handcrafted codecs. Early works used temporal architectures \cite{wu2018video, habibian2019video,golinski2020frae}, but follow-up work quickly adopted subnetworks for motion compensation and residual coding in the low latency \cite{lu2019dvc, rippel2019lvc, agustsson2020ssf, pourreza2021extending, rippel2021elfvc} and streaming setting \cite{ladune2021conditional, pourreza2021extending}. Recently, neural video codecs have adopted advanced motion compensation techniques \cite{hu2021fvc} and conditional coding, allowing them to become competitive with standard codecs in the low latency setting \cite{li2022hybrid, li2023neural}.

\subsection{Standard stereo video codecs}

Although single-view codecs achieve strong compression performance, applying them to the stereo (and more broadly multi-view) domain by independently coding each view would lead to a suboptimal linear increase in rate.
For this reason, early works in image coding extended support to stereo images by using \emph{disparity compensation} \cite{Perkins1992,Lukacs1986}. The idea is to encode one view independently, then predict the other view, for instance, with motion compensation.  Then, the difference between this prediction and the ground truth is quantized and transmitted.

Subsequent standard works, like the Multiview Video Coding (MVC) \cite{Vetro2011}, extended standard video codecs \cite{Wiegand2003,Sullivan2012} using variations of disparity compensation. In particular, the most recent standard MV-HEVC \cite{Tech2016} adopted new techniques like the coding tree unit \cite{Kim2012} to compress the disparity information. Such techniques have a few major drawbacks. First, they use many handcrafted components which cannot be optimized end-to-end. This makes it challenging to optimize reconstructions for perceptual quality or for downstream vision tasks, such as in the automotive use case.  Second, relying on explicit disparity compensation requires sequential processing of each view and therefore leads to poor parallelization across views.

\subsection{Neural stereo video codecs}

There are multiple works on neural stereo \emph{image} coding \cite{Liu2019-kt,Deng2021-ca,Lei2022-dt,Wodlinger2022-vv}. To the best of our knowledge, there is only one prior work on neural compression of stereo video, called LSVC \citep{Chen2022-xe}.  All of these works apply some form of explicit disparity compensation. As an example of a recent stereo image coding work, SASIC by \citet{Wodlinger2022-vv} uses a shared codec between left and right views but only encodes the differences in latent space between the horizontally shifted right latent and the left latent when compressing the left view.  By operating in feature space rather than pixel space, this codec allows capturing big disparities with few parameters due to the heavy spatial subsampling in the encoder. Recently, LSVC \cite{Chen2022-xe} was the first method to propose an end-to-end neural method for stereo video coding.  The main idea is to first encode the right view, then use this to conditionally encode the left view. LSVC uses a reference buffer that keeps track of the last encoded inter and intra view frames.  Using these frames, explicit feature-based motion (as originally introduced in FVC \cite{hu2021fvc}) and disparity compensation are used.  LSVC vastly outperforms the MV-HEVC standard on three common benchmarks.

% \vspace{-0.1in}
\section{Method}
% \vspace{-0.05in}
\label{sec:method}
\subsection{Redundancy and mutual information reduction}
Consider a stereo video denoted by $\{\mathbf{X}^L_t, \mathbf{X}^R_t \}_{t \in \{1 \cdots T\}}$ consisting of $T$ frames captured concurrently by the left ($L$) and right ($R$) cameras. This video contains two primary types of redundancy: (1)~temporal redundancy between consecutive frames $(\mathbf{X}^L_t$, $\mathbf{X}^L_{t+1})$ and $(\mathbf{X}^R_t$, $\mathbf{X}^R_{t+1})$; and (2)~cross-view redundancy between $\mathbf{X}^L_t$ and $\mathbf{X}^R_t$.

The performance of a video codec is significantly influenced by its ability to eliminate redundant information. For instance, temporal redundancy is commonly addressed using motion compensation techniques, where one image is aligned with another to ``reuse'' decoded information through a set of highly compressible motion vectors such as optical flows~\cite{agustsson2020ssf,lu2019dvc,sheng2022temporal} or deformable kernels~\cite{hu2021fvc,zhihao2022c2f,li2023neural}.  For stereo video compression, as illustrated in Figure~\ref{fig:teaser}, each frame at time $t$ typically needs to perform two motion compensation steps: an \emph{intra-view} step, where a prediction of the the current frame is based on the same camera view from the previous frame, and an \emph{inter-view} step, which relies on the other camera view in the current frame for prediction. Conventionally, these processes are executed in a sequential pattern~\cite{Tech2016,Chen2022-xe} which hinders the opportunities for exploiting parallel processing and leveraging specific mutual information characteristics of the stereo videos. 

\begin{figure}
    \centering
    \includegraphics[width=0.8\columnwidth]{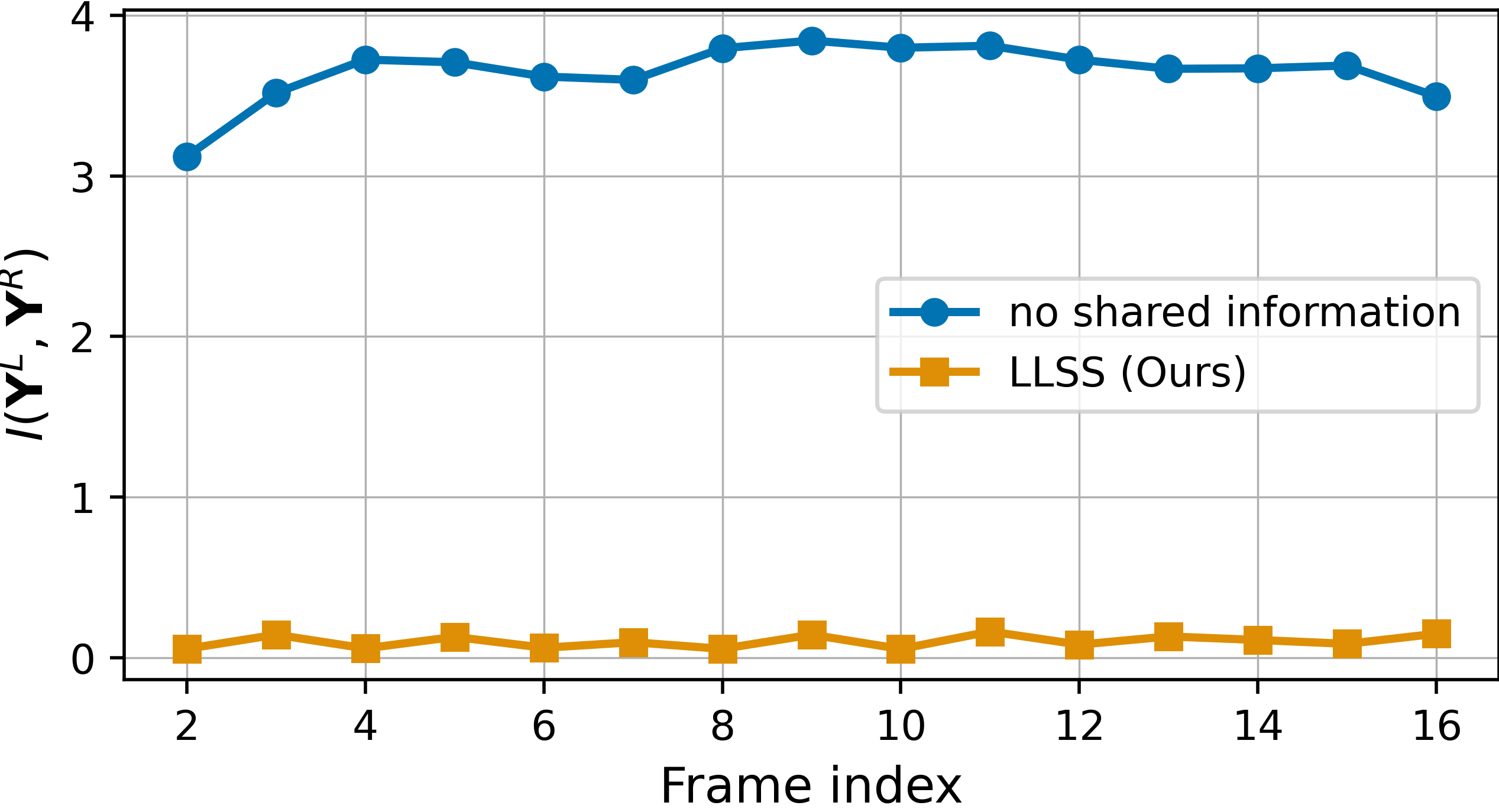}\vspace{-0.15in}
    \caption{Mutual information between cross-view motion latents. $I(\mathbf{Y}^R; \mathbf{Y}^L)=-1/2\log_2(1-\rho^2)$  for a joint Gaussian distribution with a normalized cross-correlation $\rho$.
    }
    \label{fig:cos_similarity_w_wo} \vspace{-0.25in}
\end{figure}

\begin{figure*}
\centering
  \includegraphics[width=1.0\textwidth]{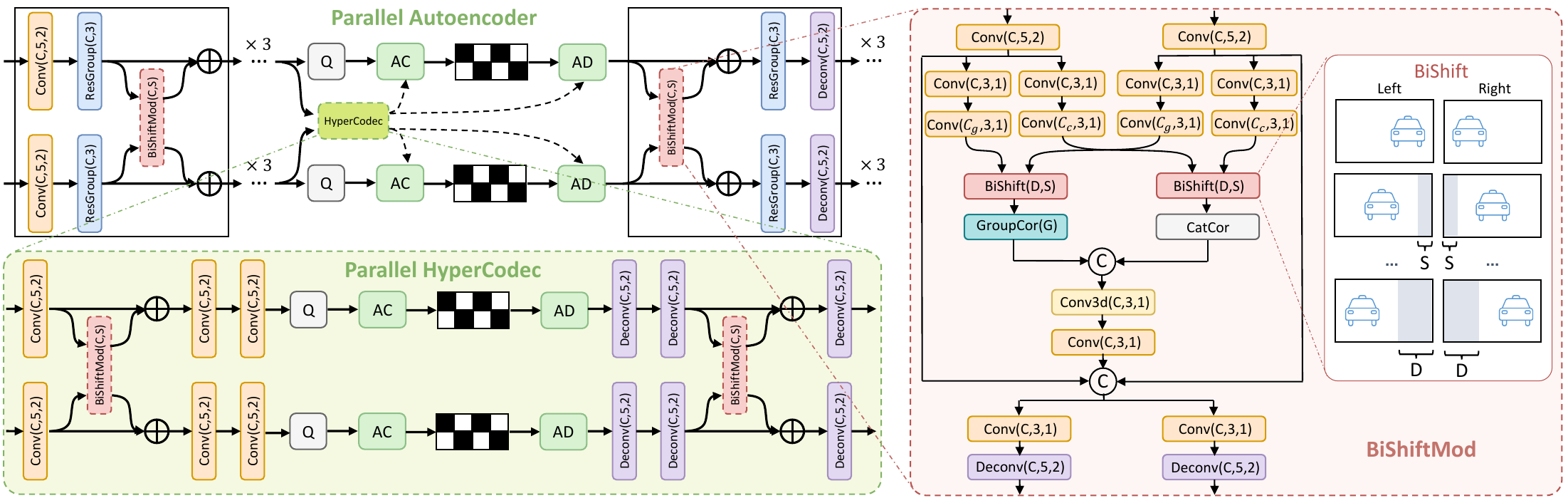}\vspace{-0.15in}
  \caption{
    The architecture of a parallel autoencoder. It contains two parallel branches to compress the left and right features at the same time. The format reads ``BlockType(channel, kernel\_size, stride)''. The Bidirectional Shift Module (BiShiftMod) is designed to learn the correlation between the left and right branches. It shifts the left and right features bidirectionally, estimating the Groupwise Correlation (GroupCor) features and Concatenation-based Correlation (CatCor) features between them. We omit activation layers for conciseness.
  }\vspace{-0.2in} 
  \label{fig:network2} 
\end{figure*} 

Stereo videos are typically rectified and highly-correlated. For instance, the disparity between two views are always in the horizontal direction~\cite{shen2022pcw,guo2019group,chang2018pyramid}. By reducing the  redundant information between two views, we could compress the stereo videos more effectively. From the rate-distortion theory~\cite{Cover2006}, the total bit rate $\mathcal{R}_L + \mathcal{R}_R$ to encode the two separate latents $\mathbf{Y}^L$ and $\mathbf{Y}^R$ generated by encoders/decoders of left and right views (with shared information)

\begin{equation}
    \mathcal{R}_L + \mathcal{R}_R \geq I(\mathbf{X}^R_t, \mathbf{X}^L_t; \mathbf{Y}^L_t, \mathbf{Y}^R_t) 
\end{equation}

\noindent where $\mathcal{R}_L, \mathcal{R}_R$ indicates the bit rate for the left and right view, respectively, and $I(\mathbf{X}^R_t, \mathbf{X}^L_t; \mathbf{Y}^L_t, \mathbf{Y}^R_t)$ indicates the mutual information between the pair of the random variables $(\mathbf{X}^R_t, \mathbf{X}^L_t)$ and $(\mathbf{Y}^L_t, \mathbf{Y}^R_t)$. It is upper bounded by 

{\footnotesize  
\begin{equation}
    {   I(\mathbf{X}^R_t, \mathbf{X}^L_t; \mathbf{Y}^L_t, \mathbf{Y}^R_t) \leq I(\mathbf{X}^R_t, \mathbf{X}^L_t; \mathbf{Y}^L_t) + I(\mathbf{X}^R_t, \mathbf{X}^L_t; \mathbf{Y}^R_t),}
\end{equation}
}
where $I(\mathbf{X}^R_t, \mathbf{X}^L_t; \mathbf{Y}^L_t)$ is indicating the mutual information~\cite{Cover2006} between joint random variable $(\mathbf{X}^R_t, \mathbf{X}^L_t)$ and random variable $\mathbf{Y}^L_t$. 
Similarly, $I(\mathbf{X}^R_t, \mathbf{X}^L_t; \mathbf{Y}^R_t)$ is the mutual information~\cite{Cover2006}  between joint random variable $(\mathbf{X}^R_t, \mathbf{X}^L_t)$ and random variable $\mathbf{Y}^R_t$. 

This causes the bit rate overhead $\mathcal{W} \geq 0$
{\footnotesize  
\begin{align}
     \mathcal{W} &= \left[I(\mathbf{X}^R_t, \mathbf{X}^L_t; \mathbf{Y}^L_t) + I(\mathbf{X}^R_t, \mathbf{X}^L_t; \mathbf{Y}^R_t)\right] - I(\mathbf{X}^R_t, \mathbf{X}^L_t; \mathbf{Y}^L_t, \mathbf{Y}^R_t) \nonumber \\
     &= I(\mathbf{Y}^L_t;\mathbf{Y}^R_t) - I(\mathbf{Y}^L_t;\mathbf{Y}^R_t | \mathbf{X}^R_t, \mathbf{X}^L_t),
\end{align}
}

\noindent comparing to the case we had a single joint encoder/decoder with the single joint latent $\mathbf{U}_t = (\mathbf{Y}^L_t, \mathbf{Y}^R_t)$ and $\mathcal{R} \geq I(\mathbf{X}^R_t, \mathbf{X}^L_t; \mathbf{Y}^L_t, \mathbf{Y}^R_t)$. $I(\mathbf{Y}^L_t;\mathbf{Y}^R_t | \mathbf{X}^R_t, \mathbf{X}^L_t)$  is a Conditional mutual information~\cite{Cover2006} of random variables $\mathbf{Y}^L_t$ and $\mathbf{Y}^R_t $ conditioned on joint random variables $(\mathbf{X}^R_t,\mathbf{X}^L_t)$.

If $I(\mathbf{Y}^L_t;\mathbf{Y}^R_t)=0$,  since $I(\mathbf{Y}^L_t;\mathbf{Y}^R_t | \mathbf{X}^R_t, \mathbf{X}^L_t) \geq 0$, the total bit rate overhead $\mathcal{W}$ has to be equal to $0$. Consequently, to reduce the total bit rate overhead $\mathcal{W}$, the network should be designed to minimize $I(\mathbf{Y}^L_t;\mathbf{Y}^R_t)$. In our architecture, we enabled this via information-sharing between the two parallel autoencoders. Specifically, we make use of two autoencoders for left and right views, and let them share information through a (learned) shifted attention module after each convolutional block.  These modules enable the information flow between the two left and right branches, helping the network to learn to reduce their redundancy, or equivalently the mutual information $I(\mathbf{Y}^L;\mathbf{Y}^R)$. Besides, the parallel design is beneficial for the parallel processing.

As shown in Figure~\ref{fig:cos_similarity_w_wo}, we observe a substantial reduction in the mutual information $I(\mathbf{Y}^L_t;\mathbf{Y}^R_t)$ between the left and right view latents when the two autoencoders are configured to share information compared to operating independently.

Given this approach of enabling more information flow, the remaining question is how to design a module that can efficiently capture and exhibit the mutual information between two branches of a codec. As inspired by the recent success of the recent state-of-the-art approach for stereo matching~\cite{shen2022pcw}, this paper introduces a Bidirectional Shift module to effective capture and transfer the mutual information between views in stereo video compression. As a learnt component, this module naturally adapts to both flow and context latent between two views. The following sections present in details the network design of our method.

\subsection{Low-Latency Neural Stereo Streaming}

Figure~\ref{fig:network} shows an overview of the architecture of LLSS. At each time step $t$, our method compresses both left and right view in a parallel manner. LLSS uses two branches, left and right, to accordingly compress  the left and right image $\mathbf{X}^L_t$ and $\mathbf{X}^R_t$ into two separated latents $\mathbf{Y}^L_t$ and $\mathbf{Y}^L_t$. The network of each branch is partially adopted from the recent feature-based video compression methods~\cite{hu2021fvc,li2021deep,li2023neural}, including the feature extractor, motion estimation, motion compensation module, and the image reconstruction module. Please find the implementation details in the supplementary materials. 

To enable inter-view information flow, LLSS introduces a novel Bidirectional Shift Module, dubbed \emph{BiShiftMod} that is inspired from the recent work from ~\citet{Wodlinger2022-vv}. This module is used to bridge the intermediate features of the network in both codec and hyper codec of the two branches.

\subsubsection{Parallel AutoEncoders}
As illustrated in Figure~\ref{fig:network}, LLSS has two pairs of parallel autoencoders: \textit{parallel motion autoencoder} and \textit{ parallel context autoencoder}. Inside each of them, there are two autoencoders running in parallel corresponding to the left and right views. The architecture of each single autoencoder is adopted from recent state-of-the-art feature-based video codec~\cite{hu2021fvc}. Due to the limited space, we would like to refer the reader to the original paper~\cite{hu2021fvc} for more details of its architecture. Briefly, each autoencoder contains a residual-based encoder to transform its input into highly compressible latent, which is then coded with the help of a hyper prior network before being decoded back to the expected output via another residual-based decoder.

To enhance the intra-view information flow, the residual autoencoder compressing the residual feature $\mathbf{R}_t=\mathbf{F}_t - \bar{\mathbf{F}}_t$, originally in FVC~\cite{hu2021fvc} is replaced by a conditional autoencoder inspired by~\cite{li2022hybrid}. 
In this conditional autoencoder, $\mathbf{F}_t$ is fed directly into the encoder, and both encoder and decoder are conditioned on the warped feature $\bar{\mathbf{F}}_t$. 
Additionally, $\bar{\mathbf{F}}_t$ is fed into hyper codec to enhance the estimation of the parameters of the prior model. 

To boost the inter-view information flow, we propose the 
``\emph{Bidirectional Shift Module}''.
In summary, this block connects the modules of encoders and decoders of the left and right branches together to enable the flow of information across views,
as illustrated in Figure~\ref{fig:network2}.
The next section details the implementation of this block.

\begin{figure*}[t]
  \centering
  \begin{subfigure}[b]{0.325\textwidth}
    \includegraphics[width=\textwidth]{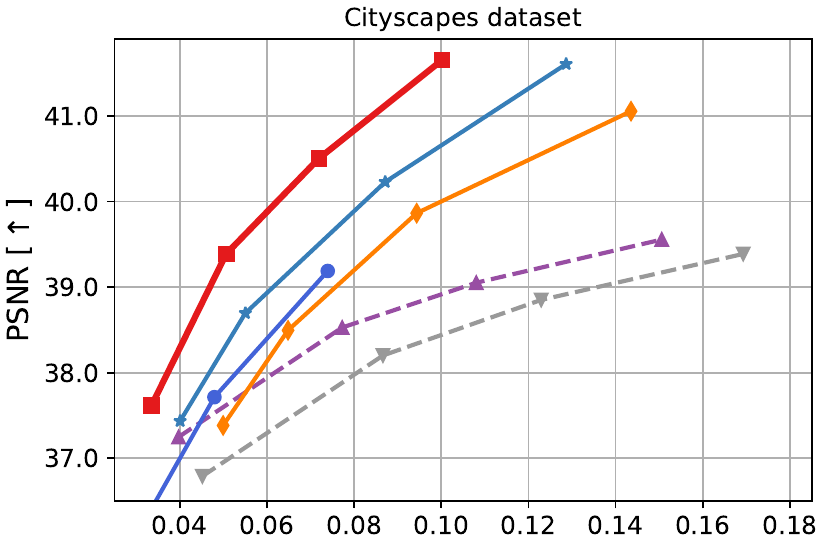}
  \end{subfigure}
  \begin{subfigure}[b]{0.31\textwidth}
    \includegraphics[width=\textwidth]{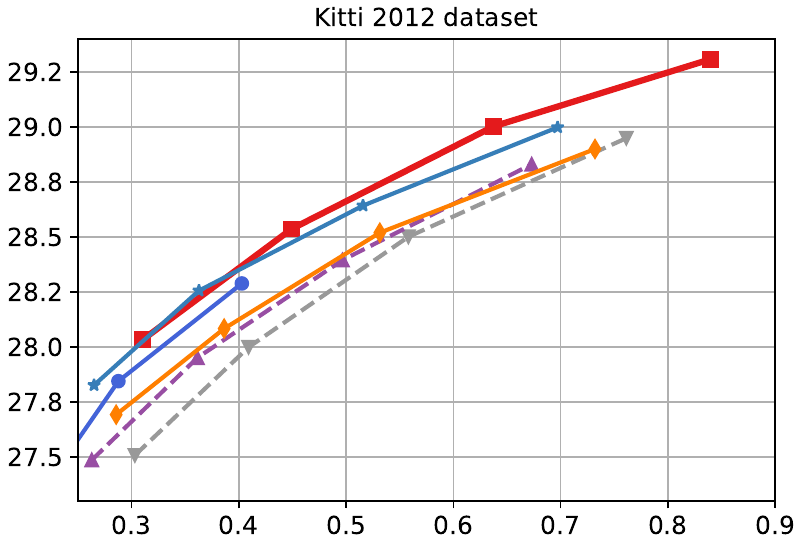}
  \end{subfigure}
  \begin{subfigure}[b]{0.31\textwidth}
    \includegraphics[width=\textwidth]{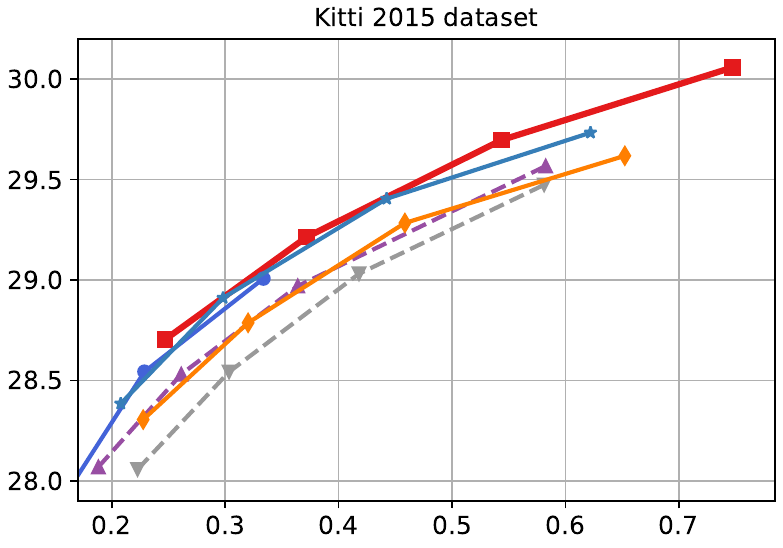}
  \end{subfigure}

  \medskip

  \begin{subfigure}[b]{0.325\textwidth}
    \includegraphics[width=\textwidth]{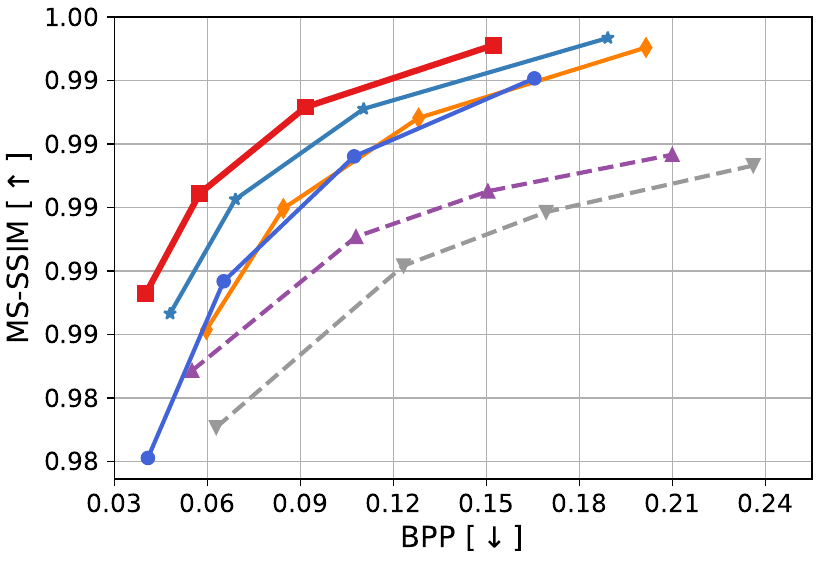}
  \end{subfigure}
  \begin{subfigure}[b]{0.31\textwidth}
    \includegraphics[width=\textwidth]{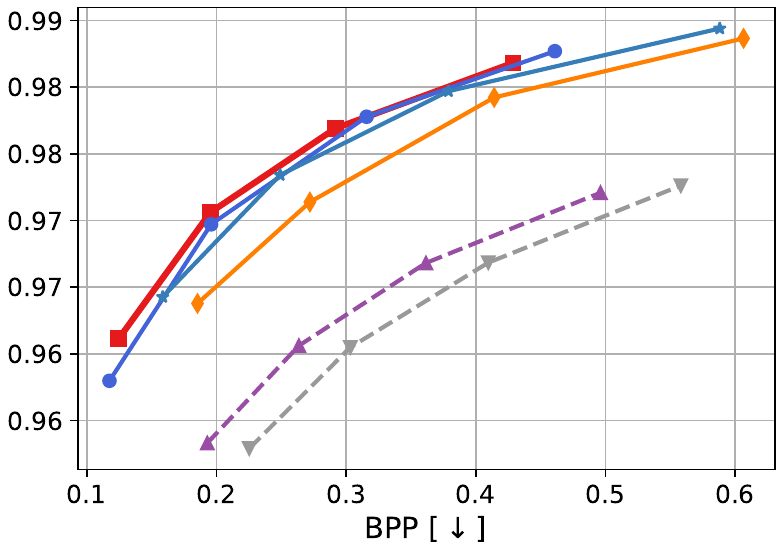}
  \end{subfigure}
  \begin{subfigure}[b]{0.315\textwidth}
    \includegraphics[width=\textwidth]{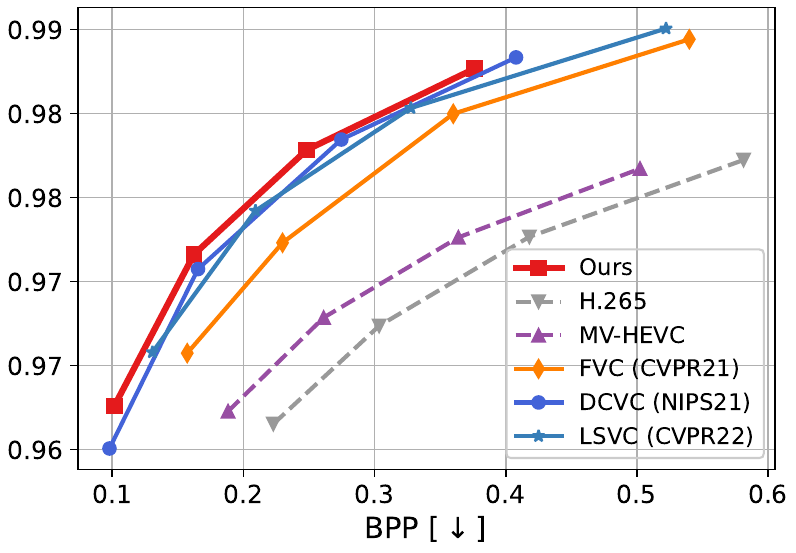}
  \end{subfigure}

  \vspace{-0.15in}
  \caption{
    Rate-distortion curves in terms of PSNR and MS-SSIM on the CityScapes~\cite{Cordts2016-de} , KITTI 2012~\cite{Geiger2012-cy} and KITTI 2015~\cite{Menze2015-rw} datasets.
  } \vspace{-0.2in}
  \label{fig:main_results}
\end{figure*}

\subsubsection{Bidirectional Shift Module}
Figure~\ref{fig:network2} shows the architecture of a Bidirectional Shift Module. It takes as input a pair of inter-view features, one from each branch of the codec, and outputs a pair of enhanced features. Inside, the inter-view features will be first transformed into a more representative form of intermediate features via a set of group-based convolutions. These intermediate features are then shifted via a module \texttt{BiShift(D,S)}. The shifted features are then passed through a set of Groupwise and Concatenation-based blocks to estimate their correlation. 

The estimated correlations are then further transformed together with the input inter-view features. These sets of transformations and correlations help capturing the redundancy between the inter-view features while feeding them into each of the left and right encoders/decoders. The encoders/decoders can then share their information efficiently and reduce the mutual information between the inter-view latents before compression.

We describe some key components of Bidirectional Shift Module in detail:
\begin{itemize}
    \item \textbf{Bidirectional shift (\texttt{BiShift(D,S))})}: horizontally shifts left feature $\mathbf{F}^L$ to the left and the right feature $\mathbf{F}^R$ to the right with a max disparity $D$ and a stride $S$.
    \item \textbf{Groupwise correlation (\texttt{GroupCor(G)})}: inspired by the stereo matching networks~\cite{shen2022pcw,guo2019group,chang2018pyramid}, measures similarity between the shifted features. It splits the features into groups and calculates the cosine distance for each group. Following~\cite{shen2022pcw,guo2019group}, the shifted features are evenly divided into $G$ groups along the channel dimension. The groupwise correlation is calculated by 
\begin{equation}
    \mathbf{V}_{gwc}(d,x,y,g) = \frac{1}{C_g/G} \langle \mathbf{F}^L_g(x, y),  \mathbf{F}^R_g(x-d, y) \rangle,
\end{equation}
where $\langle \cdot, \cdot \rangle$ and $(x, y)$ respectively indicates the inner product 
 and the pixel coordinates. $g$ and $d$ designate the index of the groups and the disparity levels. 
$\mathbf{V}_{gwc}$ is defined in $[D, H, W, G]$, where $H, W$ indicate the height and width of the feature map, respectively.

\item \textbf{Concatenation-based correlation (\texttt{CatCor})}: adopting from~\cite{shen2022pcw,guo2019group,chang2018pyramid}, it captures the similarity between the  shifted features by just concatenating them. Compared to the Groupwise correlation, it would provide more context information, thus help guide the network to learn the redundancy between the left and right branches. To get the Concatenation-based feature maps~\cite{shen2022pcw}, the shifted features are concatenated as follows
\begin{equation}
    \mathbf{V}_{concat}(d,x,y) = \mathbf{F}^L(x, y) \Vert \mathbf{F}^R(x - d, y),
\end{equation}
where $\Vert$ indicates the concatenation operator along the shifted channels. 

\end{itemize}

\subsection{Loss Function}

We optimize the entire network for the left and right views in an end-to-end manner. We adopt the typical rate-distortion loss \cite{theis2017lossy,habibian2019video} as follows
\begin{equation}
\small
     \mathcal{L} = \sum_{v\in \{L, R\}} \sum_t D(\mathbf{X}_t^v, \hat{\mathbf{X}}_t^v) + \beta\left(\mathbb{H}(\mathbf{Y}_{M,t}^v) + \mathbb{H}(\mathbf{Y}_{C,t}^v)\right),
\end{equation}
where $D(\cdot)$ indicates the distortion metric for the reconstructed frames. Depending on the training phase, it can be either the MSE or MS-SSIM loss. The superscript $v$ indicates which view is considered between Left and Right. For each view $v$ and time step $t$, $\mathbf{X}_t^v$ indicates the ground truth frame, $\hat{\mathbf{X}}_t^v$ the reconstructed frame, $\mathbf{Y}_{M,t}^v$ the quantized motion latent and $\mathbf{Y}_{C,t}^v$ the quantized context latent. $\mathbb{H}(\cdot)$ indicates the entropy function, which is proportional to the bitrate. $\beta$ indicates the hyper-parameter used to control the trade-off between the frame distortion and the rate. Note that we omit the hyper latents entropy for conciseness.

% \vspace{-0.1in}
\section{Experiments}
% \vspace{-0.05in}
\label{sec:exp}
\begin{table}[t]
\setlength{\tabcolsep}{5pt}
  \centering
  \footnotesize
  \caption{BD-rate (\%) on the CityScapes~\cite{Cordts2016-de}, KITTI 2012~\cite{Geiger2012-cy}, and KITTI 2015~\cite{Menze2015-rw} datasets. MV-HEVC~\cite{Tech2016} is set as the baseline. Lower is better, a negative number indicating bitrate \emph{savings}.}
  \label{tab:bdbr}
  \begin{tabular}{lccc}
  \toprule
     Method &CityScapes~\cite{Cordts2016-de}   &KITTI 2012~\cite{Geiger2012-cy} &KITTI 2015~\cite{Menze2015-rw}  \\ 
    \midrule
    HEVC~\cite{hevc} &33.3 &7.9 &12.7 \\ %\multirow{2}{*}{CityScape}
    FVC~\cite{hu2021fvc} &-15.6 &-2.3  &1.0   \\
    DCVC~\cite{li2021deep} &-15.2 &-13.7 &-12.3 \\
    LSVC~\cite{Chen2022-xe} &-32.7 &-17.1  &-13.4   \\
    Ours &\textbf{-50.6} &\textbf{-18.2} &\textbf{-15.8}\\
    \bottomrule
  \end{tabular} \vspace{-0.2in}
\end{table}

\noindent \textbf{Datasets.}
We make use of 4 different datasets. For training, we use the single-view Vimeo90K dataset~\cite{Xue2017-zi} for pre-training and then the stereo-camera CityScape dataset~\cite{Cordts2016-de} train set for finetuning. For evaluation we use Cityscapes test set and the stereo-video KITTI 2012~\cite{Geiger2012-cy} and 2015~\cite{Menze2015-rw} datasets.

The CityScapes~\cite{Cordts2016-de}  testing dataset comprises 1,525 30-frame stereo sequence pairs, with each containing two streams of size $2048 \times 1024$. The KITTI 2012 and 2015 testing datasets include 195 and 200 stereo sequence pairs, respectively, each containing 21 frames.
We follow LSVC \cite{Chen2022-xe} data pre-processing for the CityScape and KITTI datasets, all frames are cropped into size $1920 \times 704$ and $1216 \times 320$, respectively.

\noindent \textbf{Evaluation metrics.} We measure rate in bits-per-pixel (BPP), and assess reconstruction fidelity with the commonly used Peak Signal-to-Noise Rate (PSNR) and Multi-Scale Structural SIMilarity (MS-SSIM) \cite{wang2003msssim} metrics. To summarize the rate-distortion curve in a single number, we also report the Bj{\o}ntegaard-Delta rate (BD-rate) \cite{bjontegaard2001bdrate}, which can be interpreted as an average bitrate saving for a fixed quality compared to a reference codec. All scores are reported in the RGB color space. We evaluate our methods with a Group-of-Picture (GoP) size equal to the total sequence length i.e., 30 and 21 frames for the CityScape and KITTI datasets, respectively. For model efficiency, we report the number of parameters along with FLOP and MAC per pixel. We measure the inference GPU time using the function {\small\texttt{torch.cuda.Event()}} as well as the function  {\small\texttt{torch.cuda.synchronize()}} from the official PyTorch library~\cite{paszke2019pytorch}, while FLOPs and MACs are calculated using {\small\texttt{get\_model\_profile}} from the {\small\texttt{DeepSpeed}} library~\cite{aminabadi2022deepspeed,deepspeed}.

\begin{figure*}[ht]
\centering
  \includegraphics[width=0.32\textwidth]{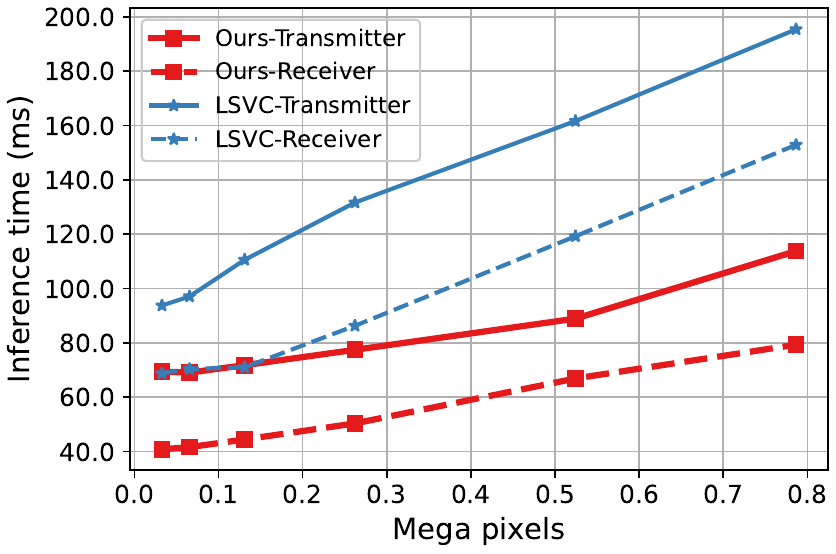}
  \includegraphics[width=0.326\textwidth]{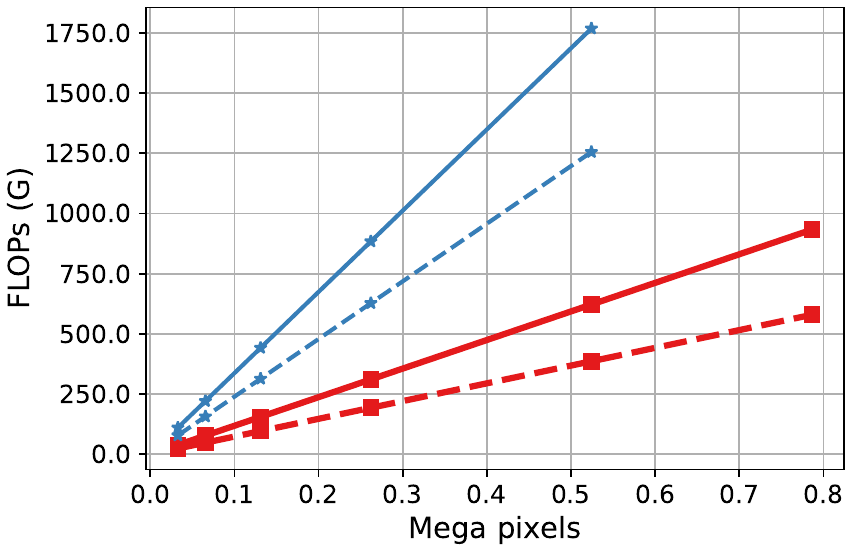}
  \includegraphics[width=0.32\textwidth]{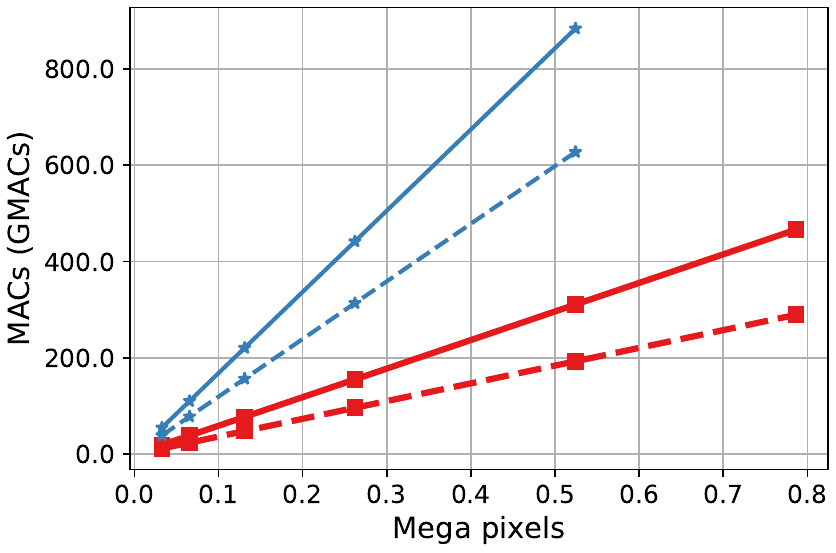} \vspace{-0.15in}
  \caption{Detailed complexity with respect to the pixel number on a single Nvidia 3080 GPU. The transmitter comprises the entire network, while the receiver is tasked solely with a subset of functions to reconstruct the frame from bits.} ~\vspace{-0.3in}
  \label{fig:complexity}
\end{figure*}

\noindent \textbf{Training details.}
We implemented our neural stereo video codec using PyTorch \cite{paszke2019pytorch}. Following a similar strategy to~\cite{Chen2022-xe}, we train our models in 3 stages: 

First, a single view version of our model (hence without the ``BiShiftMod'' modules) is randomly initialized. We train the single view model on the Vimeo90k dataset and make use of its size and diversity. We train for 2M iterations with a learning rate of $5\cdot10^{-5}$, using MSE as distortion loss. Second, the resulting pre-trained weights from the first stage are used to initialize both branches of the full stereo network. We then train the BiShiftMod modules while freezing all other modules. During this step, we train the network for 10k iterations on the CityScape dataset with a learning rate of $1 \cdot 10^{-5}$. We found this step to greatly stabilize the training process. Finally, we finetune the entire network on the CityScape dataset for 200k iterations with a learning rate of $1 \cdot 10^{-5}$. When reporting MS-SSIM performance, we use a version of our network which is further finetuned using MS-SSIM as distortion loss for an additional 100k iterations.

Across all stages, we use the Adam optimizer \cite{kingma2017adam} and train with various $\beta$ values, specifically $\left[0.0002, 0.0004, 0.0008, 0.0016, 0.0032\right]$, to  obtain rate curves. We use a batch size of 8 for the Vimeo dataset and 4 for the CityScape dataset. Additionally, we applied standard data augmentation during training. Specifically, we generated training samples by randomly cropping with size $256\times256$ for Vimeo-90k, and size $384 \times 256$ for CityScape. Our network was trained on two NVIDIA V100 GPUs for the first stage and only one for the other stages.

\noindent \textbf{Standard Baselines.} We compare our work to two standard baselines: H.265 \cite{Sullivan2012} and its multi-view extension MV-HEVC \cite{Tech2016}. We obtain the results of the standard codecs from the LSVC~\cite{Chen2022-xe} paper. For H.265, it uses the HM-16.20~\cite{hevc} implementation in the ``$\text{lowdelay}\_\text{P}\_\text{main}$'' preset on each view independently. MV-HEVC is from the HTM-16.3 implementation~\cite{mvhevc} with ``$\text{baseCfg}\_\text{2view}$'' preset.

\noindent \textbf{Learned Baselines.} The only learned stereo video codec to date is LSVC by \citet{Chen2022-xe}. Like them, we include a comparison to a single-view codec FVC by \citet{hu2021fvc}, in which feature-based warping and residual compensation were introduced and inspired LSVC architecture. We report the scores of H.265, MV-HEVC, FVC, and LSVC as recorded in \citet{Chen2022-xe}. Besides, we also compare with DCVC~\cite{li2021deep}. Since DCVC didn't release the training codes, we evaluated it without finetuning. Therefore, the comparison between our method with DCVC should be interpreted with a grain of salt.

% For the complexity analysis, we re-implemented the core operations of LSVC in PyTorch and used DeepSpeed to get the compute complexity.

\subsection{Comparison with state-of-the-art methods}

Figure~\ref{fig:main_results} and Table~\ref{tab:bdbr} respectively show the rate-distortion curves and BD-rate (with MV-HEVC as anchor) of all methods considered on the CityScapes, KITTI 2012 and 2015 test sets. Our LLSS method outperforms both learned and standard state-of-the-art methods. On the CityScape dataset, our method achieves \textbf{50.6\%} BD-rate savings compared to MV-HEVC, while LSVC only saved 32.7\%. On the KITTI 2012 and 2015 datasets, our method attains \textbf{18.2\%} and \textbf{15.8\%} BD-rate savings, respectively. 

Note that for KITTI datasets, the gap in R-D performance to LSVC has tightened. We tested our method on the KITTI datasets without finetuning, following LSVC. However, KITTI and CityScapes datasets have different data distributions due to camera settings and baselines, and image processing settings. These differences lead to significant variations in BPP and PSNR ranges when applying conventional and data-driven methods codecs. Especially all neural codecs relying on training data so tend to be less performant in this setting. Despite these challenges, our LLSS method still achieves comparable and often better results than all existing conventional and neural codecs. Compared to LSVC, our method obtains an improvement of 1.1\% and 2.4\% BD-rate gain on KITTI 2012 and KITTI 2015, respectively, while being much faster. This demonstrates the effectiveness and generalization capability of our method and its potential to further improve its performance in the future with more generalized datasets.

\begin{figure*}[ht]
\centering
  \includegraphics[width=0.32\textwidth]{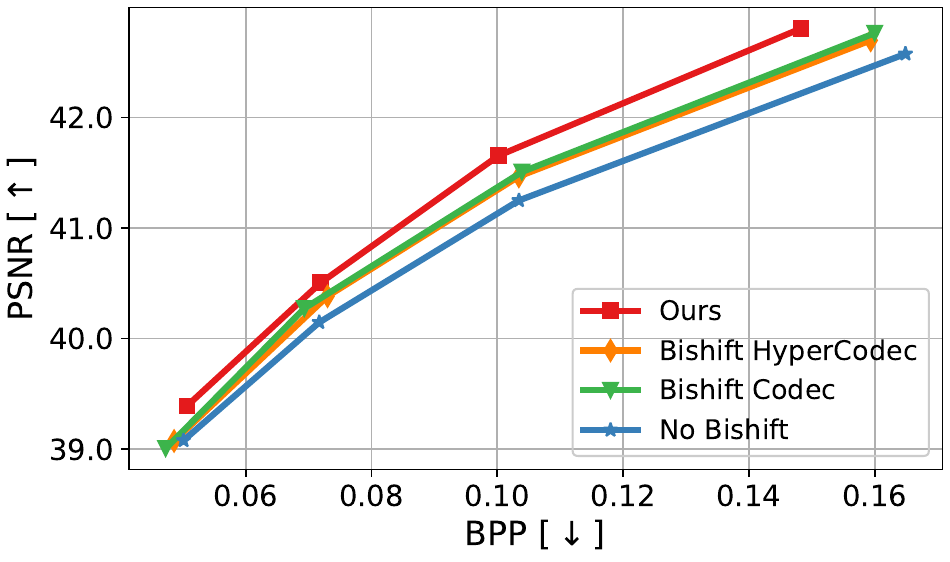}
  \includegraphics[width=0.328\textwidth]{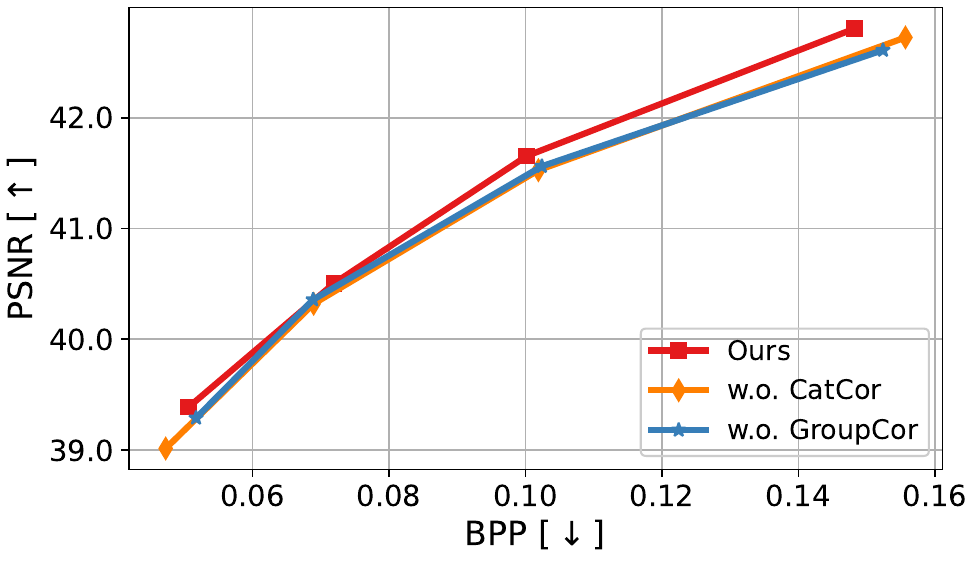}
  \includegraphics[width=0.32\textwidth]{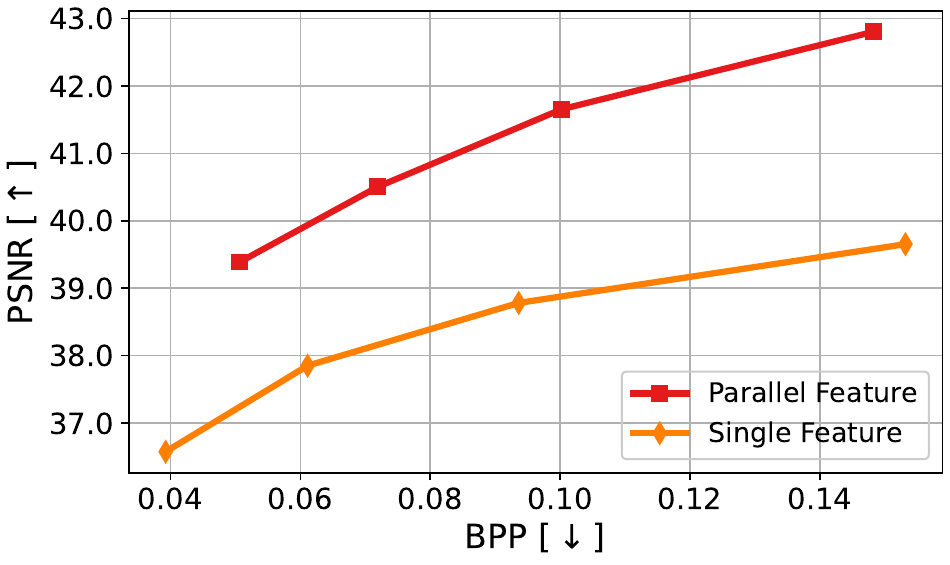}~\vspace{-0.15in}
  \caption{The effectiveness of the BiShiftMod, the components in BishiftMod, and the parallel feature on the CityScape dataset~\cite{Cordts2016-de}.} ~\vspace{-0.3in}
  \label{fig:ablation}
\end{figure*}

\subsection{Computational complexity study}

In this study, we evaluate the complexity of both the transmitter and receiver components. The transmitter encompasses the entire network, as both encoding and decoding operations are carried out in order to create the bit streams. The receiver however only encompasses the feature extractor for previous frames, the parallel motion decoding module, the motion compensation, the parallel context decoding module, and the image reconstructor. As stated in Section \ref{sec:exp}, note that LSVC does not report complexity numbers, hence we re-implemented their architecture in order to get the complexity numbers.

We reduced the complexity of our method by observing that the pixel displacement in cross-view disparity of a pair of stereo frames is simpler and more predictable than the one caused by temporal motion. When compressing temporal motion and disparity cross views, LSVC requires large and complex networks (MRC and DRC). Our method greatly simplifies this by designing an efficient BiShiftMod to align features cross-view. In the supplementary, we showed that our BiShiftMod accounts for only a small fraction of the overall computational complexity. Due to BiShiftMod, our parallel autoencoders have been designed to be more streamlined and efficient.

We examine the complexity of the transmitter and receiver in terms of inference time, FLoating-point OPerations (FLOPs), and Multiply-Add Cumulation (MACs). To investigate how these metrics perform for various video sizes, we crop the videos to the sizes including $128\times128$, $256\times128$, $256\times256$, $512\times256$, $512\times512$, and $768\times512$. Our experiments are conducted on a single Nvidia 3080 GPU, with a batch size of 1. We report the complexity for one pair of stereo P-frames and compare our method to the state-of-the-art stereo video compression approach, LSVC~\cite{Chen2022-xe}. As illustrated in Figure~\ref{fig:complexity}, our method successfully reduces computational complexity across all examined metrics. For instance, considering a pair of stereo frames of size $512\times512$, our transmitter achieves an inference time $1.7\times$ times faster than LSVC, while our receiver is $1.9\times$ times quicker. In terms of FLOPs, LSVC exhibits $2.8\times$ and $3.2\times$ times higher complexity for the transmitter and receiver, respectively. Similarly, for MACs, LSVC demonstrates $2.8\times$ and $3.3\times$ times higher complexity for the transmitter and receiver, respectively.

\subsection{Ablation study}

\textbf{Effectiveness of the BiShiftMod.} We evaluate the effectiveness of the BiShiftMod by conducting tests it on the codec and hypercodec. When training the network without BiShiftMod, we skip the second training step. Figure~\ref{fig:ablation} demonstrates that BiShiftMod significantly enhances the rate-distortion (RD) performance on the CityScape dataset. Specifically, employing BishiftMod on the codec and hypercodec lead to BD-rate savings of 7.3\% and 6.1\% compared to the configuration without BishftMod, respectively. When applied to the entire network, BishiftMod achieves 13.0\% BD-rate reduction. These improvements can be attributed to BiShiftMod's robust ability to reduce rates, highlighting the efficacy of our BiShiftMod architecture.

\noindent \textbf{Effectiveness of the BishiftMod components}. We remove the component from BishftMod, including the groupwise correlation (CroupCor) and concatenation-based correlation (CatCor). Figure~\ref{fig:ablation} shows the RD curve on the cityscape dataset. Both GroupCor and CatCor improve the results. Removing  GroupCor and CatCor results in 3.7\% and 4.3\% BD-rate increasing, respectively, which demonstrates the effectiveness of our network architecture.

\noindent \textbf{Effectiveness of parallel feature}. We concatenate the left and right features as a single feature. Figure~\ref{fig:ablation} shows that our paralleled feature streams approach achieves 62\% BD-rate saving compared to using single shared feature. Our approach to split features into views and explicitly model their cross-view redundancy is more effective than simply asking a network to perform all of these operations implicitly, which is consistent with the many other works on monocular video compression tasks that compressing frames without modeling the motion tends to spend more bits.

\begin{figure}[t]
  \centering
  \includegraphics[width=0.48\textwidth]{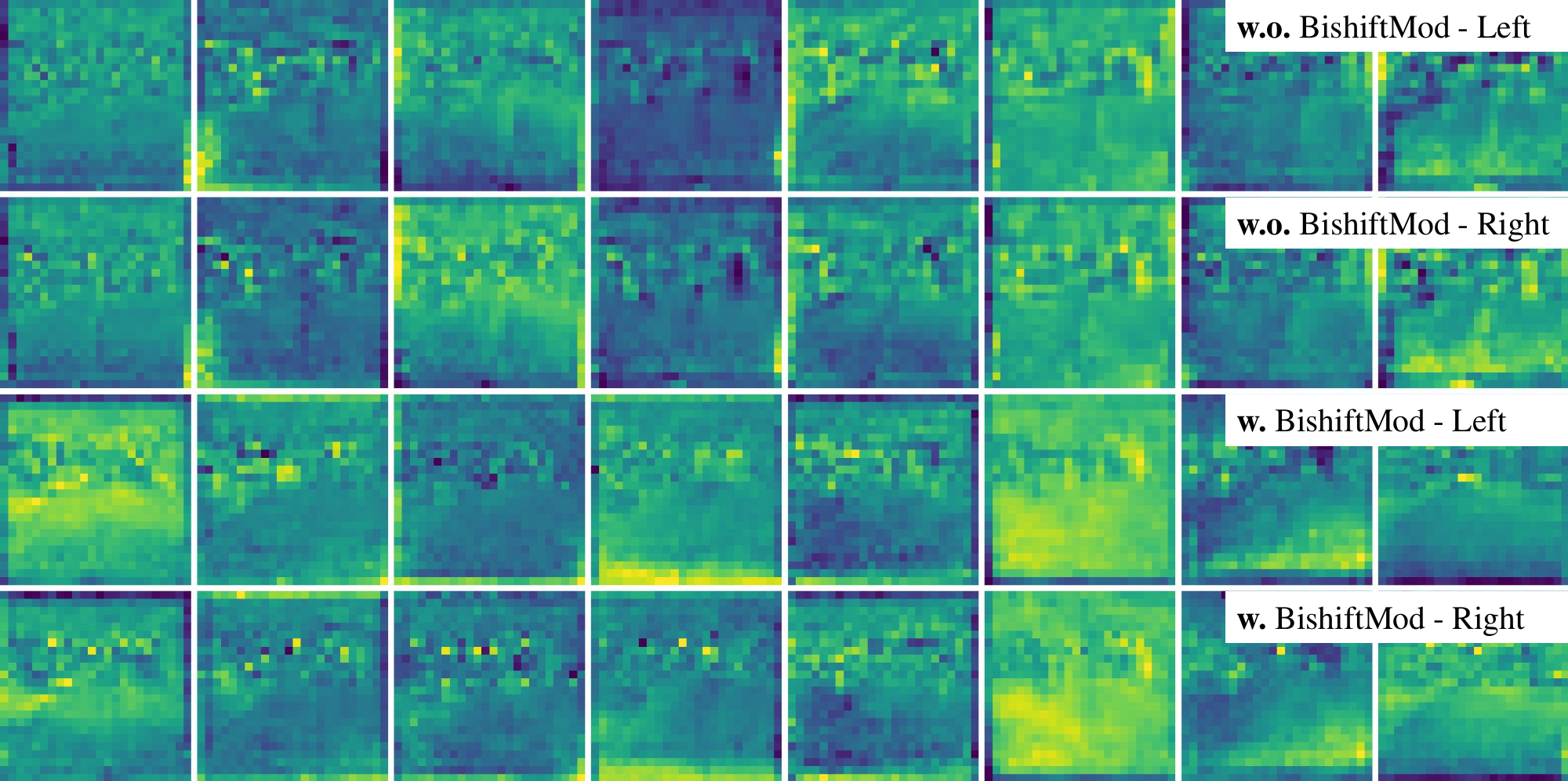}
  \caption{
    Visualization of sorted channels in the motion latents with the top-8 largest average energy. With BishiftMod, the latent features between left and right views become less alike. 
  }\vspace{-0.2in}
  \label{fig:vis_latent}
\end{figure}

\noindent \textbf{Visualization of latent features.} Figure~\ref{fig:vis_latent} displays the latent features from the left and right branches with the top-8 largest average energy. We compare the differences between the left and right branches. The first two rows are from the model without BiShiftMod, while the bottom two rows are from the model with BiShiftMod. In the absence of BiShiftMod, the latent features appear very similar. However, when BiShiftMod is present, the latent features become less alike, indicating that BiShiftMod successfully reduces redundancy between the left and right branches.

% \vspace{-0.1in}
\section{Conclusion}
% \vspace{-0.05in}
We present a low-latency neural stereo video compression method designed to simultaneously compress left and right views. We develop a bidirectional-shift compression network for this purpose. The bidirectional-shift module effectively and efficiently captures the redundancy between the left and right frames. Our experiments demonstrate that our method significantly outperforms other state-of-the-art approaches. Furthermore, the experiments show that our bidirectional-shift module and parallel autoencoders contribute to the reduced bit rates and improved frame quality.

\noindent \textbf{Acknowledgements} We thank Auke Wiggers, Reza Pourreza for their insightful feedback. Additionally, we thank Wenyu Xia for providing the voiceover in our video demo.

{
    \small
    \bibliographystyle{ieeenat_fullname}
    \bibliography{main}

\begin{thebibliography}{67}
\providecommand{\natexlab}[1]{#1}
\providecommand{\url}[1]{\texttt{#1}}
\expandafter\ifx\csname urlstyle\endcsname\relax
  \providecommand{\doi}[1]{doi: #1}\else
  \providecommand{\doi}{doi: \begingroup \urlstyle{rm}\Url}\fi

\bibitem[Agustsson et~al.(2019)Agustsson, Tschannen, Mentzer, Timofte, and
  Gool]{agustsson2019extreme}
Eirikur Agustsson, Michael Tschannen, Fabian Mentzer, Radu Timofte, and Luc~Van
  Gool.
\newblock Generative adversarial networks for extreme learned image
  compression.
\newblock In \emph{Proceedings of the IEEE/CVF International Conference on
  Computer Vision}, pages 221--231, 2019.

\bibitem[Agustsson et~al.(2020)Agustsson, Minnen, Johnston, Balle, Hwang, and
  Toderici]{agustsson2020ssf}
Eirikur Agustsson, David Minnen, Nick Johnston, Johannes Balle, Sung~Jin Hwang,
  and George Toderici.
\newblock Scale-space flow for end-to-end optimized video compression.
\newblock In \emph{Proceedings of the IEEE/CVF Conference on Computer Vision
  and Pattern Recognition}, pages 8503--8512, 2020.

\bibitem[Agustsson et~al.(2023)Agustsson, Minnen, Toderici, and
  Mentzer]{agustsson2022multirealism}
Eirikur Agustsson, David Minnen, George Toderici, and Fabian Mentzer.
\newblock Multi-realism image compression with a conditional generator.
\newblock In \emph{Proceedings of the IEEE/CVF Conference on Computer Vision
  and Pattern Recognition}, pages 22324--22333, 2023.

\bibitem[Aminabadi et~al.(2022)Aminabadi, Rajbhandari, Zhang, Awan, Li, Li,
  Zheng, Rasley, Smith, Ruwase, et~al.]{aminabadi2022deepspeed}
Reza~Yazdani Aminabadi, Samyam Rajbhandari, Minjia Zhang, Ammar~Ahmad Awan,
  Cheng Li, Du Li, Elton Zheng, Jeff Rasley, Shaden Smith, Olatunji Ruwase,
  et~al.
\newblock Deepspeed inference: Enabling efficient inference of transformer
  models at unprecedented scale.
\newblock \emph{arXiv preprint arXiv:2207.00032}, 2022.

\bibitem[Ball{\'e} et~al.(2018)Ball{\'e}, Minnen, Singh, Hwang, and
  Johnston]{balle2018variational}
Johannes Ball{\'e}, David Minnen, Saurabh Singh, Sung~Jin Hwang, and Nick
  Johnston.
\newblock Variational image compression with a scale hyperprior.
\newblock In \emph{International Conference on Learning Representations}, 2018.

\bibitem[Bjontegaard(2001)]{bjontegaard2001bdrate}
Gisle Bjontegaard.
\newblock Calculation of average psnr differences between rd-curves.
\newblock \emph{ITU SG16 Doc. VCEG-M33}, 2001.

\bibitem[Bross et~al.(2021)Bross, Wang, Ye, Liu, Chen, Sullivan, and
  Ohm]{bross2021vvc}
Benjamin Bross, Ye-Kui Wang, Yan Ye, Shan Liu, Jianle Chen, Gary~J. Sullivan,
  and Jens-Rainer Ohm.
\newblock Overview of the versatile video coding (vvc) standard and its
  applications.
\newblock \emph{IEEE Transactions on Circuits and Systems for Video
  Technology}, 31\penalty0 (10):\penalty0 3736--3764, 2021.

\bibitem[Cai et~al.(2019)Cai, Chen, Zhang, and Gao]{cai2019end}
Chunlei Cai, Li Chen, Xiaoyun Zhang, and Zhiyong Gao.
\newblock End-to-end optimized roi image compression.
\newblock \emph{IEEE Transactions on Image Processing}, 29:\penalty0
  3442--3457, 2019.

\bibitem[Chang and Chen(2018)]{chang2018pyramid}
Jia-Ren Chang and Yong-Sheng Chen.
\newblock Pyramid stereo matching network.
\newblock In \emph{Proceedings of the IEEE conference on computer vision and
  pattern recognition}, pages 5410--5418, 2018.

\bibitem[Chen et~al.(2022)Chen, Lu, Hu, Liu, Jiang, and Xu]{Chen2022-xe}
Zhenghao Chen, Guo Lu, Zhihao Hu, Shan Liu, Wei Jiang, and Dong Xu.
\newblock {LSVC}: A {Learning-Based} stereo video compression framework.
\newblock In \emph{Proceedings of the {IEEE/CVF} Conference on Computer Vision
  and Pattern Recognition}, pages 6073--6082, 2022.

\bibitem[Cordts et~al.(2016)Cordts, Omran, Ramos, Rehfeld, Enzweiler, Benenson,
  Franke, Roth, and Schiele]{Cordts2016-de}
Marius Cordts, Mohamed Omran, Sebastian Ramos, Timo Rehfeld, Markus Enzweiler,
  Rodrigo Benenson, Uwe Franke, Stefan Roth, and Bernt Schiele.
\newblock The cityscapes dataset for semantic urban scene understanding.
\newblock In \emph{Proceedings of the {IEEE} conference on computer vision and
  pattern recognition}, pages 3213--3223, 2016.

\bibitem[Cover and Thomas(2006)]{Cover2006}
Thomas~M. Cover and Joy~A. Thomas.
\newblock \emph{Elements of Information Theory}.
\newblock Wiley-Interscience, 2006.

\bibitem[DeepSpeed(2023)]{deepspeed}
DeepSpeed.
\newblock Deepspeed flops profiler.
\newblock
  \url{https://github.com/microsoft/DeepSpeed/tree/master/deepspeed/profiling/flops_profiler},
  2023.
\newblock [Online; accessed 1-May-2023].

\bibitem[Deng et~al.(2021)Deng, Yang, Yang, Xu, Liu, Feng, and
  Timofte]{Deng2021-ca}
Xin Deng, Wenzhe Yang, Ren Yang, Mai Xu, Enpeng Liu, Qianhan Feng, and Radu
  Timofte.
\newblock Deep homography for efficient stereo image compression.
\newblock In \emph{Proceedings of the {IEEE/CVF} Conference on Computer Vision
  and Pattern Recognition}, pages 1492--1501, 2021.

\bibitem[Fathima et~al.(2023)Fathima, Petersen, Sauti{\`e}re, Wiggers, and
  Pourreza]{fathima2023spatialratedistortion}
Noor Fathima, Jens Petersen, Guillaume Sauti{\`e}re, Auke Wiggers, and Reza
  Pourreza.
\newblock A neural video codec with spatial rate-distortion control.
\newblock In \emph{Proceedings of the IEEE/CVF Winter Conference on
  Applications of Computer Vision}, pages 5365--5374, 2023.

\bibitem[for Creators(2022)]{meta_quest_2022}
Meta~Quest for Creators.
\newblock Encoding immersive videos for meta quest 2.
\newblock
  \url{https://creator.oculus.com/getting-started/media-production-specifications-for-delivery-to-meta-quest-2-headsets/},
  2022.
\newblock Accessed: 2023-11-16.

\bibitem[Geiger et~al.(2012)Geiger, Lenz, and Urtasun]{Geiger2012-cy}
Andreas Geiger, Philip Lenz, and Raquel Urtasun.
\newblock Are we ready for autonomous driving? the {KITTI} vision benchmark
  suite.
\newblock In \emph{2012 {IEEE} Conference on Computer Vision and Pattern
  Recognition}, pages 3354--3361, 2012.

\bibitem[Ghouse et~al.(2023)Ghouse, Petersen, Wiggers, Xu, and
  Sautiere]{ghouse2023residual}
Noor~Fathima Ghouse, Jens Petersen, Auke Wiggers, Tianlin Xu, and Guillaume
  Sautiere.
\newblock A residual diffusion model for high perceptual quality codec
  augmentation.
\newblock \emph{arXiv preprint arXiv:2301.05489}, 2023.

\bibitem[Goliński et~al.(2020)Goliński, Pourreza, Yang, Sautière, and
  Cohen]{golinski2020frae}
Adam Goliński, Reza Pourreza, Yang Yang, Guillaume Sautière, and Taco~S.
  Cohen.
\newblock Feedback recurrent autoencoder for video compression.
\newblock \emph{ACCV}, 2020.

\bibitem[Guo et~al.(2019)Guo, Yang, Yang, Wang, and Li]{guo2019group}
Xiaoyang Guo, Kai Yang, Wukui Yang, Xiaogang Wang, and Hongsheng Li.
\newblock Group-wise correlation stereo network.
\newblock In \emph{Proceedings of the IEEE/CVF Conference on Computer Vision
  and Pattern Recognition}, pages 3273--3282, 2019.

\bibitem[Habibian et~al.(2019)Habibian, Rozendaal, Tomczak, and
  Cohen]{habibian2019video}
Amirhossein Habibian, Ties~van Rozendaal, Jakub~M Tomczak, and Taco~S Cohen.
\newblock Video compression with rate-distortion autoencoders.
\newblock In \emph{Proceedings of the IEEE/CVF International Conference on
  Computer Vision}, pages 7033--7042, 2019.

\bibitem[He et~al.(2022)He, Yang, Peng, Ma, Qin, and Wang]{he2022elic}
Dailan He, Ziming Yang, Weikun Peng, Rui Ma, Hongwei Qin, and Yan Wang.
\newblock Elic: Efficient learned image compression with unevenly grouped
  space-channel contextual adaptive coding.
\newblock \emph{2022 IEEE/CVF Conference on Computer Vision and Pattern
  Recognition (CVPR)}, 2022.

\bibitem[HEVC(2023{\natexlab{a}})]{hevc}
HEVC.
\newblock Hevc test model (hm).
\newblock \url{https://hevc.hhi.fraunhofer.de/HM-doc/}, 2023{\natexlab{a}}.
\newblock [Online; accessed 19-Apr-2023].

\bibitem[HEVC(2023{\natexlab{b}})]{mvhevc}
MV HEVC.
\newblock Multiview high efficiency video coding (mv-hevc).
\newblock \url{https://hevc.hhi.fraunhofer.de/mvhevc}, 2023{\natexlab{b}}.
\newblock [Online; accessed 19-Apr-2023].

\bibitem[Hu et~al.(2021)Hu, Lu, and Xu]{hu2021fvc}
Zhihao Hu, Guo Lu, and Dong Xu.
\newblock {FVC}: A new framework towards deep video compression in feature
  space.
\newblock In \emph{Proceedings of the {IEEE/CVF} Conference on Computer Vision
  and Pattern Recognition}, pages 1502--1511, 2021.

\bibitem[Hu et~al.(2022)Hu, Lu, Guo, Liu, Jiang, and Xu]{zhihao2022c2f}
Zhihao Hu, Guo Lu, Jinyang Guo, Shan Liu, Wei Jiang, and Dong Xu.
\newblock Coarse-to-fine deep video coding with hyperprior-guided mode
  prediction.
\newblock In \emph{Proceedings of the IEEE/CVF Conference on Computer Vision
  and Pattern Recognition}, pages 5921--5930, 2022.

\bibitem[Kim et~al.(2012)Kim, Min, Lee, Han, and Park]{Kim2012}
Il-Koo Kim, Junghye Min, Tammy Lee, Woo-Jin Han, and JeongHoon Park.
\newblock Block partitioning structure in the hevc standard.
\newblock \emph{IEEE Transactions on Circuits and Systems for Video
  Technology}, 22\penalty0 (12):\penalty0 1697--1706, 2012.

\bibitem[Kingma and Ba(2014)]{kingma2017adam}
Diederik~P Kingma and Jimmy Ba.
\newblock Adam: A method for stochastic optimization.
\newblock \emph{arXiv preprint arXiv:1412.6980}, 2014.

\bibitem[Kingma and Welling(2013)]{kingma2013autoencoding}
Diederik~P Kingma and Max Welling.
\newblock Auto-encoding variational bayes.
\newblock \emph{arXiv preprint arXiv:1312.6114}, 2013.

\bibitem[Ladune et~al.(2021)Ladune, Philippe, Hamidouche, Zhang, and
  D{\'e}forges]{ladune2021conditional}
Th{\'e}o Ladune, Pierrick Philippe, Wassim Hamidouche, Lu Zhang, and Olivier
  D{\'e}forges.
\newblock Conditional coding for flexible learned video compression.
\newblock \emph{ICLR neural compression workshop}, 2021.

\bibitem[Le et~al.(2022{\natexlab{a}})Le, Pourreza, Said, Sautiere, and
  Wiggers]{le2022gamecodec}
Hoang Le, Reza Pourreza, Amir Said, Guillaume Sautiere, and Auke Wiggers.
\newblock Gamecodec: Neural cloud gaming video codec.
\newblock In \emph{BMVC}, page 204, 2022{\natexlab{a}}.

\bibitem[Le et~al.(2022{\natexlab{b}})Le, Zhang, Said, Sautiere, Yang,
  Shrestha, Yin, Pourreza, and Wiggers]{le2022mobilecodec}
Hoang Le, Liang Zhang, Amir Said, Guillaume Sautiere, Yang Yang, Pranav
  Shrestha, Fei Yin, Reza Pourreza, and Auke Wiggers.
\newblock Mobilecodec: neural inter-frame video compression on mobile devices.
\newblock In \emph{Proceedings of the 13th ACM Multimedia Systems Conference},
  pages 324--330, 2022{\natexlab{b}}.

\bibitem[Lei et~al.(2022)Lei, Liu, Peng, Jin, Li, and Gu]{Lei2022-dt}
Jianjun Lei, Xiangrui Liu, Bo Peng, Dengchao Jin, Wanqing Li, and Jingxiao Gu.
\newblock Deep stereo image compression via {Bi-Directional} coding.
\newblock In \emph{Proceedings of the {IEEE/CVF} Conference on Computer Vision
  and Pattern Recognition}, pages 19669--19678, 2022.

\bibitem[Li et~al.(2021)Li, Li, and Lu]{li2021deep}
Jiahao Li, Bin Li, and Yan Lu.
\newblock Deep contextual video compression.
\newblock \emph{Advances in Neural Information Processing Systems}, 34, 2021.

\bibitem[Li et~al.(2022)Li, Li, and Lu]{li2022hybrid}
Jiahao Li, Bin Li, and Yan Lu.
\newblock Hybrid spatial-temporal entropy modelling for neural video
  compression.
\newblock In \emph{Proceedings of the 30th ACM International Conference on
  Multimedia}, 2022.

\bibitem[Li et~al.(2023)Li, Li, and Lu]{li2023neural}
Jiahao Li, Bin Li, and Yan Lu.
\newblock Neural video compression with diverse contexts.
\newblock In \emph{{IEEE/CVF} Conference on Computer Vision and Pattern
  Recognition, {CVPR} 2023, Vancouver, Canada, June 18-22, 2023}, 2023.

\bibitem[Liu et~al.(2019)Liu, Wang, and Urtasun]{Liu2019-kt}
Jerry Liu, Shenlong Wang, and Raquel Urtasun.
\newblock {DSIC}: Deep stereo image compression.
\newblock In \emph{Proceedings of the {IEEE/CVF} International Conference on
  Computer Vision}, pages 3136--3145, 2019.

\bibitem[Lu et~al.(2019)Lu, Ouyang, Xu, Zhang, Cai, and Gao]{lu2019dvc}
Guo Lu, Wanli Ouyang, Dong Xu, Xiaoyun Zhang, Chunlei Cai, and Zhiyong Gao.
\newblock Dvc: An end-to-end deep video compression framework.
\newblock In \emph{Proceedings of the IEEE/CVF Conference on Computer Vision
  and Pattern Recognition}, pages 11006--11015, 2019.

\bibitem[Lukacs(1986)]{Lukacs1986}
M. Lukacs.
\newblock Predictive coding of multi-viewpoint image sets.
\newblock In \emph{ICASSP '86. IEEE International Conference on Acoustics,
  Speech, and Signal Processing}, pages 521--524, 1986.

\bibitem[Mentzer et~al.(2020)Mentzer, Toderici, Tschannen, and
  Agustsson]{mentzer2020hific}
Fabian Mentzer, George Toderici, Michael Tschannen, and Eirikur Agustsson.
\newblock High-fidelity generative image compression.
\newblock \emph{Advances in Neural Information Processing Systems},
  33:\penalty0 11913--11924, 2020.

\bibitem[Menze and Geiger(2015)]{Menze2015-rw}
Moritz Menze and Andreas Geiger.
\newblock Object scene flow for autonomous vehicles.
\newblock In \emph{2015 {IEEE} Conference on Computer Vision and Pattern
  Recognition ({CVPR})}, pages 3061--3070. IEEE, 2015.

\bibitem[Minnen et~al.(2018)Minnen, Ball{\'e}, and Toderici]{minnen2018joint}
David Minnen, Johannes Ball{\'e}, and George~D Toderici.
\newblock Joint autoregressive and hierarchical priors for learned image
  compression.
\newblock \emph{Advances in neural information processing systems}, 31, 2018.

\bibitem[Muckley et~al.(2023)Muckley, El-Nouby, Ullrich, J{\'e}gou, and
  Verbeek]{muckley2023improving}
Matthew~J Muckley, Alaaeldin El-Nouby, Karen Ullrich, Herv{\'e} J{\'e}gou, and
  Jakob Verbeek.
\newblock Improving statistical fidelity for neural image compression with
  implicit local likelihood models.
\newblock \emph{arXiv preprint arXiv:2301.11189}, 2023.

\bibitem[Paszke et~al.(2019)Paszke, Gross, Massa, Lerer, Bradbury, Chanan,
  Killeen, Lin, Gimelshein, Antiga, Desmaison, Kopf, Yang, DeVito, Raison,
  Tejani, Chilamkurthy, Steiner, Fang, Bai, and Chintala]{paszke2019pytorch}
Adam Paszke, Sam Gross, Francisco Massa, Adam Lerer, James Bradbury, Gregory
  Chanan, Trevor Killeen, Zeming Lin, Natalia Gimelshein, Luca Antiga, Alban
  Desmaison, Andreas Kopf, Edward Yang, Zachary DeVito, Martin Raison, Alykhan
  Tejani, Sasank Chilamkurthy, Benoit Steiner, Lu Fang, Junjie Bai, and Soumith
  Chintala.
\newblock Pytorch: An imperative style, high-performance deep learning library.
\newblock In \emph{Advances in Neural Information Processing Systems 32}, pages
  8024--8035. Curran Associates, Inc., 2019.

\bibitem[Perkins(1992)]{Perkins1992}
M.G. Perkins.
\newblock Data compression of stereopairs.
\newblock \emph{IEEE Transactions on Communications}, 40\penalty0 (4):\penalty0
  684--696, 1992.

\bibitem[Pourreza and Cohen(2021)]{pourreza2021extending}
Reza Pourreza and Taco~S Cohen.
\newblock Extending neural p-frame codecs for b-frame coding.
\newblock \emph{Proceedings of the IEEE/CVF International Conference on
  Computer Vision}, pages 6680--6689, 2021.

\bibitem[Pourreza et~al.(2023)Pourreza, Le, Said, Sauti\`ere, and
  Wiggers]{pourreza2023boosting}
Reza Pourreza, Hoang Le, Amir Said, Guillaume Sauti\`ere, and Auke Wiggers.
\newblock Boosting neural video codecs by exploiting hierarchical redundancy.
\newblock In \emph{Proceedings of the IEEE/CVF Winter Conference on
  Applications of Computer Vision (WACV)}, pages 5355--5364, 2023.

\bibitem[Rippel et~al.(2019)Rippel, Nair, Lew, Branson, Anderson, and
  Bourdev]{rippel2019lvc}
Oren Rippel, Sanjay Nair, Carissa Lew, Steve Branson, Alexander~G. Anderson,
  and Lubomir Bourdev.
\newblock Learned video compression.
\newblock In \emph{Proceedings of the IEEE/CVF International Conference on
  Computer Vision (ICCV)}, 2019.

\bibitem[Rippel et~al.(2021)Rippel, Anderson, Tatwawadi, Nair, Lytle, and
  Bourdev]{rippel2021elfvc}
Oren Rippel, Alexander~G Anderson, Kedar Tatwawadi, Sanjay Nair, Craig Lytle,
  and Lubomir Bourdev.
\newblock Elf-vc: Efficient learned flexible-rate video coding.
\newblock \emph{Proceedings of the IEEE/CVF International Conference on
  Computer Vision}, pages 14479--14488, 2021.

\bibitem[Shen et~al.(2022)Shen, Dai, Song, Rao, Zhou, and Zhang]{shen2022pcw}
Zhelun Shen, Yuchao Dai, Xibin Song, Zhibo Rao, Dingfu Zhou, and Liangjun
  Zhang.
\newblock Pcw-net: Pyramid combination and warping cost volume for stereo
  matching.
\newblock In \emph{Computer Vision--ECCV 2022: 17th European Conference, Tel
  Aviv, Israel, October 23--27, 2022, Proceedings, Part XXXII}, pages 280--297.
  Springer, 2022.

\bibitem[Sheng et~al.(2022)Sheng, Li, Li, Li, Liu, and Lu]{sheng2022temporal}
Xihua Sheng, Jiahao Li, Bin Li, Li Li, Dong Liu, and Yan Lu.
\newblock Temporal context mining for learned video compression.
\newblock \emph{IEEE Transactions on Multimedia}, 2022.

\bibitem[Str{\"u}mpler et~al.(2021)Str{\"u}mpler, Postels, Yang, Van~Gool, and
  Tombari]{strumpler2021implicit}
Yannick Str{\"u}mpler, Janis Postels, Ren Yang, Luc Van~Gool, and Federico
  Tombari.
\newblock Implicit neural representations for image compression.
\newblock \emph{arXiv preprint arXiv:2112.04267}, 2021.

\bibitem[Sullivan et~al.(2012)Sullivan, Ohm, Han, and Wiegand]{Sullivan2012}
Gary~J. Sullivan, Jens-Rainer Ohm, Woo-Jin Han, and Thomas Wiegand.
\newblock Overview of the high efficiency video coding (hevc) standard.
\newblock \emph{IEEE Transactions on Circuits and Systems for Video
  Technology}, 22\penalty0 (12):\penalty0 1649--1668, 2012.

\bibitem[Tech et~al.(2016)Tech, Chen, Müller, Ohm, Vetro, and Wang]{Tech2016}
Gerhard Tech, Ying Chen, Karsten Müller, Jens-Rainer Ohm, Anthony Vetro, and
  Ye-Kui Wang.
\newblock Overview of the multiview and 3d extensions of high efficiency video
  coding.
\newblock \emph{IEEE Transactions on Circuits and Systems for Video
  Technology}, 26\penalty0 (1):\penalty0 35--49, 2016.

\bibitem[Theis et~al.(2017)Theis, Shi, Cunningham, and
  Husz{\'a}r]{theis2017lossy}
Lucas Theis, Wenzhe Shi, Andrew Cunningham, and Ferenc Husz{\'a}r.
\newblock Lossy image compression with compressive autoencoders.
\newblock \emph{ICLR}, 2017.

\bibitem[Toderici et~al.(2017)Toderici, Vincent, Johnston, Jin~Hwang, Minnen,
  Shor, and Covell]{toderici2017full}
George Toderici, Damien Vincent, Nick Johnston, Sung Jin~Hwang, David Minnen,
  Joel Shor, and Michele Covell.
\newblock Full resolution image compression with recurrent neural networks.
\newblock In \emph{Proceedings of the IEEE conference on Computer Vision and
  Pattern Recognition}, pages 5306--5314, 2017.

\bibitem[van Rozendaal et~al.(2021{\natexlab{a}})van Rozendaal, Brehmer, Zhang,
  Pourreza, and Cohen]{rozendaal2021instance}
Ties van Rozendaal, Johann Brehmer, Yunfan Zhang, Reza Pourreza, and Taco~S
  Cohen.
\newblock Instance-adaptive video compression: Improving neural codecs by
  training on the test set.
\newblock \emph{arXiv preprint arXiv:2111.10302}, 2021{\natexlab{a}}.

\bibitem[van Rozendaal et~al.(2021{\natexlab{b}})van Rozendaal, Huijben, and
  Cohen]{rozendaal2021overfitting}
Ties van Rozendaal, Iris~AM Huijben, and Taco~S Cohen.
\newblock Overfitting for fun and profit: Instance-adaptive data compression.
\newblock \emph{arXiv preprint arXiv:2101.08687}, 2021{\natexlab{b}}.

\bibitem[Vetro et~al.(2011)Vetro, Wiegand, and Sullivan]{Vetro2011}
Anthony Vetro, Thomas Wiegand, and Gary~J. Sullivan.
\newblock Overview of the stereo and multiview video coding extensions of the
  h.264/mpeg-4 avc standard.
\newblock \emph{Proceedings of the IEEE}, 99\penalty0 (4):\penalty0 626--642,
  2011.

\bibitem[Wang et~al.(2003)Wang, Simoncelli, and Bovik]{wang2003msssim}
Z. Wang, E.P. Simoncelli, and A.C. Bovik.
\newblock Multiscale structural similarity for image quality assessment.
\newblock In \emph{The Thrity-Seventh Asilomar Conference on Signals, Systems
  and Computers, 2003}, pages 1398--1402 Vol.2, 2003.

\bibitem[Wiegand et~al.(2003{\natexlab{a}})Wiegand, Sullivan, Bjontegaard, and
  Luthra]{Wiegand2003}
T. Wiegand, G.J. Sullivan, G. Bjontegaard, and A. Luthra.
\newblock Overview of the h.264/avc video coding standard.
\newblock \emph{IEEE Transactions on Circuits and Systems for Video
  Technology}, 13\penalty0 (7):\penalty0 560--576, 2003{\natexlab{a}}.

\bibitem[Wiegand et~al.(2003{\natexlab{b}})Wiegand, Sullivan, Bjontegaard, and
  Luthra]{wiegand2003overview}
Thomas Wiegand, Gary~J Sullivan, Gisle Bjontegaard, and Ajay Luthra.
\newblock Overview of the h. 264/avc video coding standard.
\newblock \emph{IEEE Transactions on circuits and systems for video
  technology}, 13\penalty0 (7):\penalty0 560--576, 2003{\natexlab{b}}.

\bibitem[W{\"o}dlinger et~al.(2022)W{\"o}dlinger, Kotera, Xu, and
  Sablatnig]{Wodlinger2022-vv}
Matthias W{\"o}dlinger, Jan Kotera, Jan Xu, and Robert Sablatnig.
\newblock {SASIC}: Stereo image compression with latent shifts and stereo
  attention.
\newblock In \emph{Proceedings of the {IEEE/CVF} Conference on Computer Vision
  and Pattern Recognition}, pages 661--670, 2022.

\bibitem[Wu et~al.(2018)Wu, Singhal, and Krähenbühl]{wu2018video}
Chao-Yuan Wu, Nayan Singhal, and Philipp Krähenbühl.
\newblock Video compression through image interpolation.
\newblock \emph{Proceedings of the European conference on computer vision
  (ECCV)}, pages 416--431, 2018.

\bibitem[Xue et~al.(2019)Xue, Chen, Wu, Wei, and Freeman]{Xue2017-zi}
Tianfan Xue, Baian Chen, Jiajun Wu, Donglai Wei, and William~T Freeman.
\newblock Video enhancement with task-oriented flow.
\newblock \emph{International Journal of Computer Vision}, 127:\penalty0
  1106--1125, 2019.

\bibitem[Yang et~al.(2021)Yang, Van~Gool, and Timofte]{yang2021perceptual}
Ren Yang, Luc Van~Gool, and Radu Timofte.
\newblock Perceptual learned video compression with recurrent conditional
  {GAN}.
\newblock \emph{arXiv preprint arXiv:2109.03082}, 2021.

\bibitem[Zhang et~al.(2021)Zhang, van Rozendaal, Brehmer, Nagel, and
  Cohen]{zhang2021implicit}
Yunfan Zhang, Ties van Rozendaal, Johann Brehmer, Markus Nagel, and Taco Cohen.
\newblock Implicit neural video compression.
\newblock \emph{arXiv preprint arXiv:2112.11312}, 2021.

\end{thebibliography}


\begin{thebibliography}{8}
\providecommand{\natexlab}[1]{#1}
\providecommand{\url}[1]{\texttt{#1}}
\expandafter\ifx\csname urlstyle\endcsname\relax
  \providecommand{\doi}[1]{doi: #1}\else
  \providecommand{\doi}{doi: \begingroup \urlstyle{rm}\Url}\fi

\bibitem[Ball{\'e} et~al.(2018)Ball{\'e}, Minnen, Singh, Hwang, and
  Johnston]{balle2018variational}
Johannes Ball{\'e}, David Minnen, Saurabh Singh, Sung~Jin Hwang, and Nick
  Johnston.
\newblock Variational image compression with a scale hyperprior.
\newblock In \emph{International Conference on Learning Representations}, 2018.

\bibitem[Chen et~al.(2022)Chen, Lu, Hu, Liu, Jiang, and Xu]{Chen2022-xe}
Zhenghao Chen, Guo Lu, Zhihao Hu, Shan Liu, Wei Jiang, and Dong Xu.
\newblock {LSVC}: A {Learning-Based} stereo video compression framework.
\newblock In \emph{Proceedings of the {IEEE/CVF} Conference on Computer Vision
  and Pattern Recognition}, pages 6073--6082, 2022.

\bibitem[Cordts et~al.(2016)Cordts, Omran, Ramos, Rehfeld, Enzweiler, Benenson,
  Franke, Roth, and Schiele]{Cordts2016-de}
Marius Cordts, Mohamed Omran, Sebastian Ramos, Timo Rehfeld, Markus Enzweiler,
  Rodrigo Benenson, Uwe Franke, Stefan Roth, and Bernt Schiele.
\newblock The cityscapes dataset for semantic urban scene understanding.
\newblock In \emph{Proceedings of the {IEEE} conference on computer vision and
  pattern recognition}, pages 3213--3223, 2016.

\bibitem[Geiger et~al.(2012)Geiger, Lenz, and Urtasun]{Geiger2012-cy}
Andreas Geiger, Philip Lenz, and Raquel Urtasun.
\newblock Are we ready for autonomous driving? the {KITTI} vision benchmark
  suite.
\newblock In \emph{2012 {IEEE} Conference on Computer Vision and Pattern
  Recognition}, pages 3354--3361, 2012.

\bibitem[Hu et~al.(2021)Hu, Lu, and Xu]{hu2021fvc}
Zhihao Hu, Guo Lu, and Dong Xu.
\newblock {FVC}: A new framework towards deep video compression in feature
  space.
\newblock In \emph{Proceedings of the {IEEE/CVF} Conference on Computer Vision
  and Pattern Recognition}, pages 1502--1511, 2021.

\bibitem[Le et~al.(2022)Le, Zhang, Said, Sautiere, Yang, Shrestha, Yin,
  Pourreza, and Wiggers]{Le2022-ra}
Hoang Le, Liang Zhang, Amir Said, Guillaume Sautiere, Yang Yang, Pranav
  Shrestha, Fei Yin, Reza Pourreza, and Auke Wiggers.
\newblock {MobileCodec}: neural inter-frame video compression on mobile
  devices.
\newblock In \emph{Proceedings of the 13th {ACM} Multimedia Systems
  Conference}, pages 324--330, New York, NY, USA, 2022. Association for
  Computing Machinery.

\bibitem[Menze and Geiger(2015)]{Menze2015-rw}
Moritz Menze and Andreas Geiger.
\newblock Object scene flow for autonomous vehicles.
\newblock In \emph{2015 {IEEE} Conference on Computer Vision and Pattern
  Recognition ({CVPR})}, pages 3061--3070. IEEE, 2015.

\bibitem[Misra(2019)]{misra2019mish}
Diganta Misra.
\newblock Mish: A self regularized non-monotonic activation function.
\newblock \emph{arXiv preprint arXiv:1908.08681}, 2019.

\end{thebibliography}
}

% % Arxiv
% \clearpage
% \appendix

% \section{Model Configurations}
% \input{supp/model}

% \section{Datasets Configurations}
% \input{supp/training}

% \section{Detailed Coding Complexity}
% \input{supp/more_results}

% \section{Discussion}
% \input{cvpr2024/supp/limit}

\end{document}

\maketitle

\renewcommand{\thefootnote}{\fnsymbol{footnote}}
\footnotetext[1]{\hspace{-0.1in}Corresponding author} 
\footnotetext[2]{\hspace{-0.1in}Qualcomm AI Research is an initiative of Qualcomm Technologies, Inc.} 
\renewcommand*{\thefootnote}{\arabic{footnote}}

\vspace{-0.3in}
\section{Model Configurations}
% \vspace{-0.1in}
In this section, we provide more details on the configuration for each module of our network.

\begin{figure}[t]
\centering
\vspace{-0.1in}
  \includegraphics[width=1.\columnwidth]{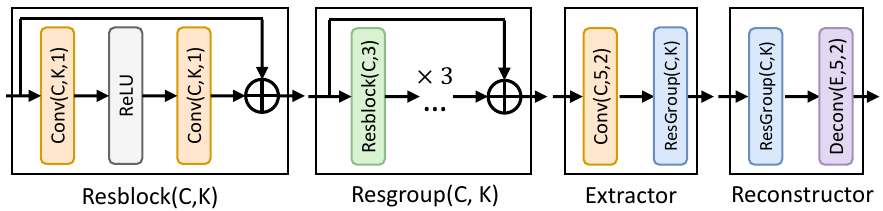}\vspace{-0.1in}
  \caption{Feature extractor and the image reconstructor.}
  \label{fig:ext_recon} \vspace{-0.2in}
\end{figure}

\noindent \textbf{Feature extractor and image reconstructor.} Figure~\ref{fig:ext_recon} shows the architecture of the feature extractor and the image reconstructor. Following FVC~\cite{hu2021fvc} and LSVC~\cite{Chen2022-xe}, the feature extractor extracts the features from the RGB input images,  which are subsequently compressed within the network. 
The extraction process involves downsampling the image using strided convolutions,  followed by a stack of residual convolution blocks. The image reconstructor synthesizes the final RGB images from the reconstructed features.  It shares a similar architecture to the feature extractor, consisting of one residual convolution block and strided transposed convolutional layers. The weights of the feature extractor and image reconstructor are shared between the left and right branches of the network. Both are including 3 ResGroup modules with a channel number of $C=64$, a kernel size of 3, and a stride of 1.

\begin{figure}[t]
\centering
\vspace{-0.1in}
  \includegraphics[width=1.\columnwidth]{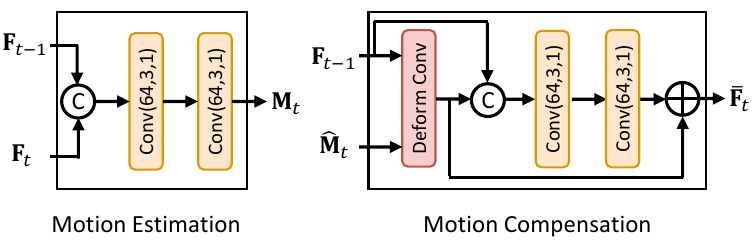}\vspace{-0.1in}
  \caption{The structure of the motion estimation module and the motion compensation module.} \vspace{-0.3in}
  \label{fig:motion}
\end{figure}

\noindent \textbf{Motion estimation and motion compensation.} Figure~\ref{fig:motion} illustrates the network architectures for the motion estimation and motion compensation modules. Following FVC~\cite{hu2021fvc}, our motion estimation module contains two convolutional layers. This module takes the features $\mathbf{F}_t$ and $\mathbf{\hat{F}}_{t-1}$  as inputs and estimates the offset vectors $\mathbf{M}_{t}$ between them. These offset vectors are then quantized, encoded, and sent to a decoder side where they will be reconstructed to $\mathbf{\hat{M}}_{t}$. The coding process is performed with the commonly used hyperprior-based network~\cite{balle2018variational}.

The motion compensation module aims to generate the current feature maps by warping $\mathbf{\hat{F}}_{t-1}$ using $\mathbf{\hat{M}}_{t}$ through a deformable convolution. Our network uses two convolutions as well as a skip connection to fuse the warped features with the previous feature $\mathbf{\hat{F}}_{t-1}$. We set the group number as 8 in the deformable convolutional layer for the motion compensation. Motion estimation module and motion compensation module contain two convolutional layers with a channel number of $C=64$, a kernel size of 3, and a stride of 1. We use \texttt{ReLU} as our activation function.

\noindent \textbf{Parallel Motion Autoencoder.} Figure 4 in the main paper shows the architecture of the parallel motion autoencoder. We set the number of channels $C=64$ for the parallel motion autoencoder. For its parallel HyperCodec, the number of channels is set to $C=128$. The \texttt{ResGroup} layers utilize a kernel size of 3 and a stride of 1. The \texttt{conv} layers, primarily designed for feature downsampling, incorporate a kernel size of 5 along with a stride of 2. Conversely, the \texttt{deconv} layers, aiming to upsample the feature, employ a kernel size of 5 and a stride of 2 as well. \texttt{ReLU} is adopted as our activation function. 

\noindent \textbf{Parallel Context Autoencoder.} The context autoencoder also uses the architecture shown in Figure 4 of the main paper. Different from the parallel motion autoencoder, we set the  number of channels as $C=128$ in the parallel context autoencoder. The channel number of its parallel HyperCodec is set to $C=128$. To match the channel number after concatenation layers, we use one convolutional layer with a channel number of 128, a kernel size of 3, and a stride of 1. We use \texttt{ReLU} for the activation function.

\noindent \textbf{Bidirectional Shift Module.} Figure 4 in the main paper shows the architecture of the bidirectional shift module. We set the channel number in bidirectional shift modules as $C=32$, $C_g=32$, and $C_c=12$. The group number of the GroupCor module is set as $G=4$. Since the max disparity for the KITTI datasets~\cite{Geiger2012-cy,Menze2015-rw} is 192, we set the max shift distance $D=192/2^{Scale - 1}$, where $Scale$ indicates the downscale factor of the input feature maps with respect to the source image. Similar to the max disparity, we set the shift stride $S=max(1, 8/2^{Scale - 1})$. The BiShiftMod component downsamples input features using \texttt{conv} layers with a kernel size of 5 and a stride of 2. Conversely, it upsamples the output features through \texttt{deconv} layers, also featuring a kernel size of 5 and a stride of 2. The remaining \texttt{conv} layers have a kernel size of 3 and a stride of 1. We use \texttt{Mish}~\cite{misra2019mish} as the activation function.

% \section{Training and Evaluation}
\vspace{-0.05in}
\section{Datasets Configurations}
We utilized the official script provided by the authors of LSVC~\cite{Chen2022-xe} to pre-process the CityScapes~\cite{Cordts2016-de} and KITTI~\cite{Geiger2012-cy,Menze2015-rw} datasets, ensuring a fair comparison.

\textbf{KITTI 2012 and 2015 datasets.} We selected the ``testing'' subset for our experiments. In accordance with LSVC~\cite{Chen2022-xe}, we excluded videos containing fewer than 21 frames. Specifically, videos ``000127" and ``000182" from the KITTI 2012 dataset, and videos ``000026" and ``000167" from the KITTI 2015 dataset were removed. We then cropped the videos to a size of $1216 \times 320$ from the top left, employing \texttt{ffmpeg} and using the "420p" pixel format.

\textbf{CityScapes datasets.} Following the approach of LSVC~\cite{Chen2022-xe}, we cropped 128 pixels from the left and 64 pixels from the bottom to eliminate artifacts resulting from rectification. Additionally, we removed 256 pixels from the bottom to exclude the ego-vehicle area. We applied \texttt{ffmpeg} with the "420p" pixel format for cropping. After the cropping, we obtained videos with sizes of $1920 \times 704$.

\vspace{-0.05in}
\section{Detailed Coding Complexity}
\begin{table}[t]
\setlength{\tabcolsep}{6pt}
  \centering
  \small
  \caption{Complexity for each component of our network for input size $512\times512$.}
  \label{tab:complexity_component}
  \begin{tabular}{lccc}
  \toprule
     Component &MACs(G)  &FLOPs(G) &Params(M) \\ 
    \midrule
    Feature extractor &59.2 &118.7 &0.23 \\ 
    Image reconstructor &42.4 &84.8 &0.23  \\
    Motion estimation &14.5 &29.1  &0.22   \\
    Motion compensation &25.4 &50.8 &0.40 \\
    Motion autoencoder &32.0 &64.0  &10.93   \\
    \hspace{0.15in}└─ BiShiftMods &2.9 &5.8 &1.67 \\
    Context autoencoder &137.3 &275.1 &23.54  \\
    \hspace{0.15in}└─ BiShiftMods &3.3 &6.6 &2.00 \\

    \bottomrule
  \end{tabular}
  \vspace{-0.2in}
\end{table}

In this study, we further examine the complexity of each component within our network. Table~\ref{tab:complexity_component} presents the complexity in terms of MACs, FLOPs, and the number of parameters (Params). The majority of calculations are concentrated on the context autoencoder, as it is designed to capture detailed context information following motion compensation. Consequently, we set a larger channel number of 128. Additionally, we report the complexity of the BiShiftMods in both the motion autoencoder and the context autoencoder. Our BiShiftMod accounts for only a small fraction of the overall computational complexity.

\vspace{-0.05in}
\section{Discussion}
Figure 6 in the main paper shows that our methods achieve both smaller operational complexity and faster inference time compared to the competitive LSVC~\cite{Chen2022-xe} method. Even when the difference in complexity is small given low resolution inputs, our method still reduces the inference time significantly compared to LSVC. This advantage is thanks to our novel architecture that is optimized for parallel processing. We note that this the inference time is still for benchmarking purpose and the current design is not aimed for deployment yet. In the future, it is possible to further optimize our network for practical deployment such as following the approach in the recent practical codec~\cite{Le2022-ra}.

\vspace{-0.05in}
{
    \small
    \bibliographystyle{ieeenat_fullname}
    \bibliography{main}
}